\documentclass[11pt, a4paper, goog]{google}

\usepackage[authoryear, sort&compress, round]{natbib}
\newcommand{\canvas}{\textsc{CANVAS}}
\usepackage{amsmath}
\usepackage{times}
\usepackage{latexsym}
\usepackage{xcolor}
\usepackage{tcolorbox}
\usepackage{amssymb}
\usepackage[table]{xcolor}

\usepackage{xcolor}
\usepackage{pifont}
\DeclareMathOperator*{\\argmax}{argmax}

\newcommand{\cmark}{\textcolor{green!60!black}{\ding{51}}}
\newcommand{\xmark}{\textcolor{red!70!black}{\ding{55}}}
\usepackage[T1]{fontenc}
\tcbuselibrary{breakable}
\usepackage[utf8]{inputenc}
\usepackage[ruled,vlined]{algorithm2e}

\uselogo{} 

\title{\textbf{CANVAS: \textbf{C}ontinuity-\textbf{A}ware \textbf{N}arratives via \textbf{V}isual \textbf{A}gentic \textbf{S}toryboarding}
}

\usepackage{pifont}
\usepackage{multirow}
\usepackage[table]{xcolor} 

\correspondingauthor{imondal@umd.edu, yiwensong@google.com, yalesong@google.com \\ *This work was done when Ishani was a Student Researcher at Google.}

\reportnumber{0001} 

\renewcommand{\today}{}

\author[1]{Ishani Mondal\textsuperscript{*}}
\author[2]{Yiwen Song}
\author[2]{Mihir Parmar}
\author[2]{Palash Goyal}
\author[1]{Jordan Boyd-Graber}
\author[2]{Tomas Pfister}
\author[2]{Yale Song}

\affil[1]{University of Maryland, College Park}
\affil[2]{Google} 

\begin{abstract}
Long-form visual storytelling requires maintaining continuity across shots, including consistent characters, stable environments, and smooth scene transitions. While existing generative models can produce strong individual frames, they fail to preserve such continuity, leading to appearance changes, inconsistent backgrounds, and abrupt scene shifts.
We introduce CANVAS (\textbf{C}ontinuity-\textbf{A}ware \textbf{N}arratives via \textbf{V}isual \textbf{A}gentic \textbf{S}toryboarding), a multi-agent framework that explicitly plans visual continuity in multi-shot narratives. CANVAS enforces coherence through character continuity, persistent background anchors, and location-aware scene planning for smooth transitions within the same setting 
We evaluate CANVAS on two storyboard generation benchmarks \textbf{ST-BENCH} and \textbf{ViStoryBench} 
and introduce a new challenging benchmark \textbf{HardContinuityBench} 
for long-range narrative consistency. CANVAS consistently outperforms the best-performing baseline, improving background continuity by 21.6\%, character consistency by 9.6\% and props consistency by 7.6\%.
\\ \\
Project Website: \url{https://ishani-mondal.github.io/canvas-project-page/}
\end{abstract}

\begin{document}

\maketitle

\section{Introduction}


\begin{figure*}[!t]
    \centering
    \includegraphics[width=\linewidth]{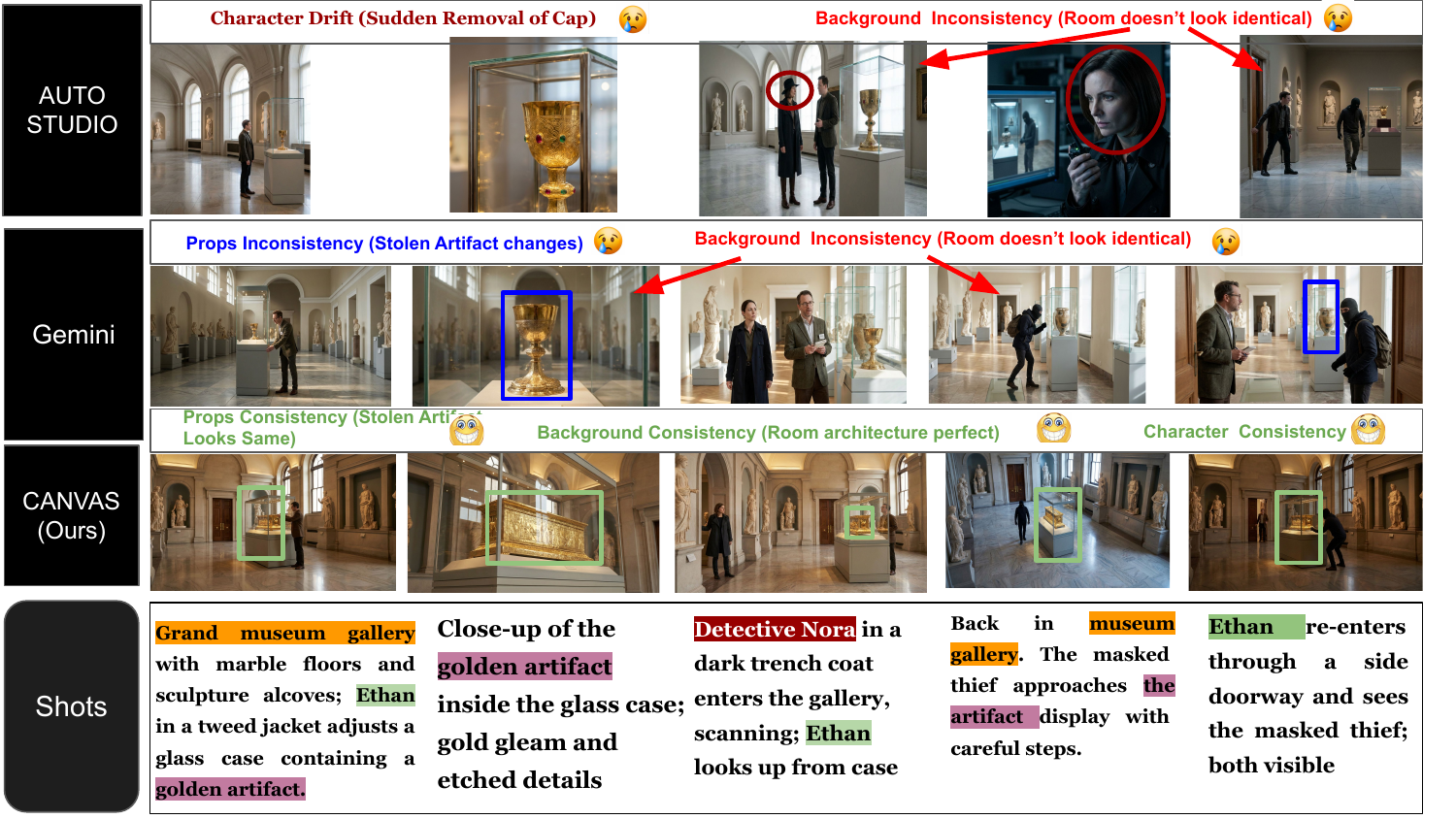}
    \caption{
Visual storyboards generated by AutoStudio \citep{cheng2025autostudiocraftingconsistentsubjects}, Gemini-3.1-pro, and CANVAS on a multi-shot museum heist story from our HardContinuityBench. Prior approaches exhibit character appearance drift, unstable scene geometry across consecutive shots, and logical inconsistencies such as artifact state changes, room layout drift across non-consecutive shots. In contrast, $\canvas$ preserves character identity, stable room layout, artifact state consistency, and object persistence across shots. We use similar colors in the shot descriptions highlighting that those should maintain identical appearance throughout the entire storyline. }
    \label{fig:teaserfig}
    \vspace{-1.3mm}
\end{figure*}

Long-form visual storytelling requires generating a sequence of images that together form a coherent narrative world \citep{li2026loststoriesconsistencybugs, natangelo2025narrativecontinuitytestconceptual}. Unlike single-image generation, storyboard generation must preserve long-range consistency across multiple shots: characters should retain stable visual identities, environments should remain consistent when locations reappear, and objects should evolve logically as narrative events unfold.
A natural way to enforce such coherence is through \textit{world-state modeling for visual generation} \citep{xiaoworldmem, World}. In this formulation, the generation process maintains an explicit representation of the narrative world—including characters, environments, and object states—that evolves over time. Image generation at each step is conditioned on this evolving narrative representation, enabling the system to maintain long-range visual consistency while faithfully reflecting narrative state changes.

Most existing approaches generate story frames \emph{sequentially}, conditioning each new image on previously generated frames or reference images \citep{dinkevich2025story2boardtrainingfreeapproachexpressive, mao2026storyitertrainingfreeiterativeparadigm, cheng2024theatergen, cheng2025autostudiocraftingconsistentsubjects, zhou2024storydiffusionconsistentselfattentionlongrange}. 
Auto-regressive methods generate frames one-by-one from earlier frames, while reference-based approaches maintain subject identity through \textbf{latent anchor or diffusion-based techniques} using a small set of reference images (Table~\ref{tab:method_comparison}). More recent iterative methods (e.g., Story-Iter \citep{mao2026storyitertrainingfreeiterativeparadigm}) refine frames by repeatedly conditioning on previously generated story frames to improve global visual similarity.
However, these methods primarily enforce short-range consistency and do not explicitly model the evolving narrative world. As a result, inconsistencies accumulate over time, leading to common failure modes such as character appearance drift, unstable environments when locations reappear, and incorrect object states after narrative events (Figure~\ref{fig:teaserfig}). 
These issues arise because story world itself is never explicitly represented. In coherent narratives, characters maintain identity unless the plot changes their appearance, environments preserve spatial structure when revisited, and objects follow causal state transitions (e.g., present $\rightarrow$ stolen $\rightarrow$ absent). 

We formulate storyboard generation as \textbf{explicit world-state modeling}, maintaining structured representations of characters, locations, and object states as the narrative evolves. We introduce $\canvas$, a multi-agent framework that enforces continuity through \textbf{world-state transitions} via iterative \textit{planning, generation, and memory updates} (Section~\ref{sec:methodology}). 
The system first constructs a global continuity plan, then generates each shot by retrieving visual anchors from memory or initializing new ones when needed (e.g., rendering a fireplace in the first Santa Claus scene in Fig~\ref{fig:shot_comparison} (a) enables smooth transitions to later shot (b), whereas without this, the background must change abruptly to accommodate future events, as seen from (c) to (d)).
Memory is updated after every generation of each frame, ensuring consistency across long-range narrative dependencies.

Another challenge in studying narrative consistency is evaluation. Existing storyboard benchmarks \citep{zhuang2025vistorybenchcomprehensivebenchmarksuite, zhang2025storymemmultishotlongvideo} rarely require long-range reasoning: recurring locations are uncommon, character appearance changes are limited, and object state transitions seldom influence future scenes. Hence, they barely capture subtle forms of narrative drift that occur in longer sequences. To address this limitation, we introduce \textbf{HardContinuityBench} (Section~\ref{sec:experimentalsetup}), a benchmark specifically designed to stress-test long-range narrative continuity (Details in Appendix~\ref{app:hardcontinuityBench_justification}).

We evaluate \textbf{\canvas} against several state-of-the-art story visualization baselines on multiple storyboard generation benchmarks like ViStoryBench \citep{zhuang2025vistorybenchcomprehensivebenchmarksuite}, ST-Bench \citep{zhang2025storymemmultishotlongvideo} and our new benchmark \textbf{HardContinuityBench}.  
We propose \textbf{ContinuityEval}, a VLM-based fine-grained metric on assessing character, background and props continuity.
$\canvas$ consistently outperforms the best-performing baseline on ContinuityEval, improving background continuity by 21.6\%, character consistency by 9.6\% and props consistency by 7.6\%  (Section~\ref{sec:mainresults}). 
Besides, $\canvas$ outperforms all the existing training-free storyboard generation methods on standard ViStoryBench metrics (Table~\ref{tab:vistorybench_results}) and video generation performances on standard FilmEval metrics \citep{huang2025filmasterbridgingcinematicprinciples} (Table~\ref{tab:video_eval_results}).

Our contributions are threefold: 
\begin{enumerate}
    \item We introduce \textbf{$\canvas$}, a training-free multi-agent framework that models storyboard generation as explicit world-state tracking through global planning, memory-guided retrieval, and sequential memory updates.
    \item We propose \textbf{HardContinuityBench}, a benchmark designed to stress-test long-range narrative consistency, capturing subtle drift in character identity, environment structure, and object state transitions.
    \item Extensive experiments show that $\canvas$ significantly improves character, background, and prop consistency across multiple storyboard benchmarks.
\end{enumerate}

\begin{table*}[t]
\centering
\begin{tabular}{lccccc}
\toprule
\textbf{Method} & 
\textbf{Char} & 
\textbf{BG} & 
\textbf{Future} & 
\textbf{Multi-} & 
\textbf{Training} \\
& \textbf{Consist} & 
\textbf{Consist} & 
\textbf{Cond} & 
\textbf{Agent} & 
\textbf{Free} \\
\midrule

StoryGen \citep{zhang2025stagestoryboardanchoredgenerationcinematic}  & \textbf{\cmark} & \textbf{\xmark} & \textbf{\xmark} & \textbf{\xmark} & \textbf{\xmark} \\

StoryDiffusion \citep{zhou2024storydiffusionconsistentselfattentionlongrange} & \textbf{\cmark} & \textbf{\xmark} & \textbf{\xmark} & \textbf{\xmark} & \textbf{\xmark} \\

Story2Board \citep{dinkevich2025story2boardtrainingfreeapproachexpressive} & \textbf{\cmark} & \textbf{\xmark} & \textbf{\xmark} & \textbf{\xmark} & \textbf{\cmark} \\

Story-Iter \citep{mao2026storyitertrainingfreeiterativeparadigm} & \textbf{\cmark} & \textbf{Partial} & \textbf{\xmark} & \textbf{\xmark} & \textbf{\cmark} \\

AutoStudio \citep{cheng2025autostudiocraftingconsistentsubjects} & \textbf{\cmark} & \textbf{\xmark} & \textbf{\xmark} & \textbf{\cmark} & \textbf{\cmark} \\

\rowcolor{green!15}
\textbf{$\canvas$} & \textbf{\cmark} & \textbf{\cmark} & \textbf{\cmark} & \textbf{\cmark} & \textbf{\cmark} \\

\bottomrule
\end{tabular}

\caption{Comparison of existing Story Visualization methods. 
Most prior methods focus on preserving character consistency across frames using reference images or attention mechanisms. 
However, these methods do not explicitly maintain recurring location representations or future-conditioned scene planning. 
Our multi-agent training-free approach tackles these challenges.}
\label{tab:method_comparison}
\end{table*}

\section{Related Work}

\begin{figure*}[!t]
    \centering
    \includegraphics[width=\linewidth]{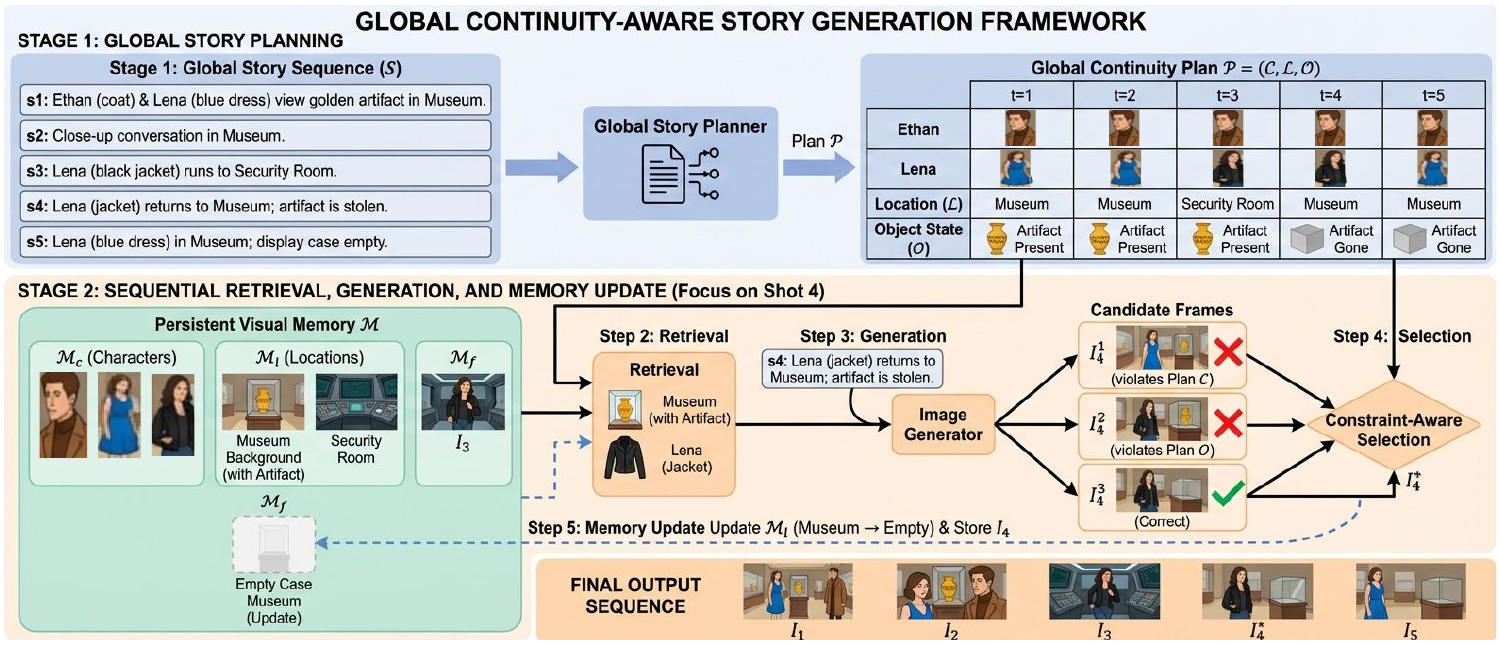}
    \caption{
\textbf{Overview of $\canvas$ Pipeline.}
The framework generates a sequence of visually coherent cinematic keyframes from a story script through global planning, sequential generation, and persistent visual memory. 
}
    \label{fig:method}
\end{figure*}

\subsection{Visual Story Generation from Text}
Early work on visual storytelling focused on generating sequences of images from textual narratives using generative models. One representative approach is StoryGen \citep{Liu_2024_CVPR}, which introduces an autoregressive diffusion-based framework for open-ended visual storytelling. The model generates each frame conditioned on the text description and preceding image-caption pairs using a vision–language context module integrated into Stable Diffusion.
During generation, features extracted from earlier frames are injected into the diffusion denoising process, allowing the model to maintain some degree of visual continuity across frames.

While StoryGen improves over independent text-to-image generation by conditioning on previous frames, its architecture primarily focuses on appearance-level consistency rather than explicit modeling of narrative state. Because generation is conditioned only on preceding frames, the system lacks a structured representation of characters, locations, or object states. Consequently, errors can accumulate over long sequences, and recurring environments or props are not guaranteed to remain stable across distant shots. In contrast, CANVAS introduces explicit world-state modeling with persistent character, background, and prop anchors, enabling reasoning about how entities evolve across the story.

Another line of work explores architectural modifications to diffusion models to improve cross-frame consistency. StoryDiffusion \citep{zhou2024storydiffusionconsistentselfattentionlongrange} introduces a consistent self-attention mechanism that injects reference tokens from previous frames into the attention computation of the diffusion model. This mechanism allows the model to maintain subject identity across generated frames while preserving the controllability of text prompts.
 However, StoryDiffusion \citep{zhou2024storydiffusionconsistentselfattentionlongrange}  focuses primarily on subject identity preservation, particularly character appearance and attire. The method does not attempt to model narrative structure, scene persistence, or object state evolution across the story. Moreover, because consistency is enforced through attention correlations rather than explicit memory structures, the model lacks a mechanism for reasoning about long-range narrative constraints, such as ensuring that objects change state only when story specifies such events.

More recent work explores training-free storyboard generation pipelines. Story2Board \citep{dinkevich2025story2boardtrainingfreeapproachexpressive} proposes a lightweight consistency mechanism combining latent panel anchoring and reciprocal attention value mixing to maintain subject identity across storyboard panels.
The system uses a language model to convert a story into panel-level prompts and then generates panels sequentially using diffusion models.

Although Story2Board \citep{dinkevich2025story2boardtrainingfreeapproachexpressive} improves panel-level coherence and allows dynamic pose and spatial variation, its design prioritizes visual expressiveness and layout diversity rather than strict narrative continuity. The framework does not maintain persistent representations of characters, locations, or props across panels, and therefore cannot explicitly enforce logical consistency when scenes reappear later in the story.

In contrast, CANVAS models story generation as an evolving narrative world state, enabling persistent tracking of characters, environments, and objects throughout the story timeline.

\subsection{Multi-Turn Image Generation and Interactive Story Creation}
Another research direction focuses on interactive image generation, where images are generated iteratively through multi-turn user instructions.

AutoStudio \citep{cheng2025autostudiocraftingconsistentsubjects} proposes a multi-agent framework for interactive image generation, combining large language models with diffusion models to maintain subject consistency across multiple turns. The architecture consists of a subject manager that interprets user instructions, a layout generator that predicts bounding boxes for subjects, a supervisor that refines layouts, and a diffusion-based image generator that produces the final image.
This design allows AutoStudio to maintain consistency for multiple subjects while responding to user instructions such as adding objects or modifying scene elements.
Despite these strengths, AutoStudio primarily addresses interactive editing and layout control rather than long-form narrative consistency. The system focuses on maintaining subject identity within short interactive sessions and does not explicitly model persistent environments or object states across a structured story. As a result, the framework does not address challenges such as maintaining stable environments when scenes reappear or ensuring that object states evolve logically according to narrative events.

Similarly, TheaterGen \citep{cheng2024theatergen} introduces a framework for multi-turn image generation in which a large language model acts as a ``screenwriter'' that generates a structured prompt book describing characters and layouts.
Character prompts are used to generate reference images during a rehearsal stage, and these references guide the final image generation during the diffusion process.
The key idea in TheaterGen is character management through prompt planning, which improves semantic consistency in multi-turn generation tasks.
However, TheaterGen primarily focuses on character-level consistency, and its architecture generates subjects individually before merging them into the final image. This design can lead to unnatural interactions between characters or inconsistencies in spatial relationships. Moreover, the framework does not maintain persistent representations of environments or objects across a story timeline.

Compared to these approaches, CANVAS focuses on story-level coherence rather than interactive editing, introducing explicit mechanisms to maintain background continuity, object persistence, and logical state transitions across narrative events.

\subsection{Memory-Based Narrative Generation}
Several recent works attempt to address long-range consistency using memory mechanisms.
StoryMem \citep{zhang2025storymemmultishotlongvideo} reformulates visual storytelling as iterative shot synthesis conditioned on a memory bank of keyframes from previously generated shots. The framework injects these keyframes into a single-shot video diffusion model using a Memory-to-Video architecture, enabling the model to preserve visual context across shots.
StoryMem improves cross-shot consistency by storing selected keyframes and using them as conditioning signals during generation.
However, the stored memory primarily functions as visual context for conditioning, rather than as an explicit representation of narrative entities. The system does not separately model characters, environments, and objects or reason about how these entities evolve across the story. As a result, the memory acts as a passive reference rather than an interpretable narrative representation.

Very recently, VideoMemory \citep{zhou2026videomemoryconsistentvideogeneration} introduces an entity-centric dynamic memory bank that stores representations of characters, props, and backgrounds. A multi-agent system retrieves these representations before generating each shot and updates the memory after generation.This architecture improves long-range entity consistency by maintaining persistent visual descriptors of story entities.
Nevertheless, VideoMemory primarily focuses on video-level entity persistence rather than storyboard-level reasoning about scene structure. The system retrieves stored representations but does not explicitly plan narrative continuity or enforce constraints such as maintaining stable environment geometry across recurring scenes. Furthermore, generation remains largely conditioned on visual descriptors rather than structured narrative reasoning.

Against Narrative Weaver \citep{yao2026narrativeweavercontrollablelongrange}, OneStory \citep{an2025onestorycoherentmultishotvideo}, and HoloCine \citep{meng2025holocineholisticgenerationcinematic} which bake long-range context or planning into the generator, CANVAS’s advantage is its architecture-agnostic, training-free deployment. However, these training-based methods report strong long-range coherence and could serve as upper-bound references.

\section{$\canvas$}
\label{sec:methodology}
Visual storyboard generation requires maintaining long-range consistency across characters, environments, and objects throughout a sequence of shots. However, existing approaches often generate frames independently, leading to character appearance drift, unstable environments when locations reappear, and incorrect object states after narrative events. To address these challenges, we propose \textbf{$\canvas$}, a framework that enforces narrative consistency through three components: \textit{global story planning} to model character, location, and object evolution; \textit{memory-guided sequential generation} to retrieve relevant visual anchors across shots; and \textit{QA-based candidate selection} to ensure that generated frames satisfy continuity constraints (Algorithm~\ref{algo:canvas}).
We illustrate the pipeline with a running example in Fig.~\ref{fig:method}. The story follows Ethan and Lena across five shots. They first examine a golden artifact in a museum gallery, later move to a security room where Lena changes into a security jacket, and eventually return to the gallery where the artifact is stolen. In the final shot, Lena reappears in her original blue dress while the display case remains empty. This example highlights the need to preserve character identity, maintain stable environments across recurring locations, and correctly update object states as the narrative unfolds.

Formally, the narrative is represented as a sequence of shot descriptions $S=\{s_1,\ldots,s_T\}$, and the goal is to generate a sequence of storyboard images $I=\{I_1,\ldots,I_T\}$ that faithfully depict each shot while remaining consistent with previously generated frames. $\canvas$ achieves this through global story planning followed by sequential generation with retrieval and memory updates.

\subsection{Global Story Planning}
\label{subsec:planning}
Independent frame generation cannot ensure narrative consistency across long sequences because the model lacks knowledge about how characters, environments, and objects evolve over time. The planning stage therefore constructs a global representation of these narrative dependencies before image generation begins.

The input to the planner is a sequence of shot descriptions $S=\{s_1,\ldots,s_T\}$. It produces a global continuity plan denoted by $\mathcal{P}=(\mathcal{C},\mathcal{L},\mathcal{O})$ with character appearance states $\mathcal{C}$, location assignments $\mathcal{L}$, and object state transitions across the story $\mathcal{O}$ (Prompt in Table~\ref{tab:prop_state_planning_prompt}). 
The plan $\mathcal{P}$ constrains how characters, environments, and objects evolve throughout the narrative and provides structured guidance for the next sequential generation stage (Section~\ref{subsec:generation}).

\paragraph{Character.}
Characters must maintain consistent visual identities---face, hairstyle, and clothing---unless the narrative dictates a change. To prevent identity drift, the planner defines a mapping $\mathcal{C}(c,t)\rightarrow k$ (Prompt in Table~\ref{tab:character_appearance_single}), specifying the appearance state $k$ for character $c$ at shot $t$. In our running example, while Ethan remains in a trench coat, Lena's timeline tracks her shift from a blue dress ($s_1, s_2$) to a security jacket ($s_3, s_4$) and back ($s_5$). These explicit constraints guide the generation stage in retrieving and updating character appearance anchors to ensure long-range coherence. 

\paragraph{Location.}
Stories frequently revisit locations, but without explicit modeling, generators often reconstruct environments inconsistently. To ensure spatial consistency, our planner clusters shots by location identity $l$ using the mapping $\mathcal{L}(s_t)\rightarrow l$ (Clustering details in Table~\ref{tab:location_clustering_prompt}) 
It further reasons about narrative events to pre-arrange key props that must persist across future scenes (Prompt in Table~\ref{tab:future_prop_background_prompt}). In our museum example, assigning consistent identities to shots $(s_1, s_2, s_4, s_5)$ allows the system to reuse stored background anchors, preserving the overall scene structure when location reappears.

\paragraph{Object State.}
Objects often change state, altering the visual environment when moved or removed. Without explicit tracking, generators may produce ``ghost'' objects or outdated scene elements. The planner models this evolution through the mapping $\mathcal{O}(o,t)\rightarrow\text{state}$ (Prompt in Table~\ref{tab:prop_state_planning_prompt}), specifying an object $o$'s status at shot $t$. In our museum example, the golden artifact's transition from ``present'' ($s_1 - s_3$) to ``stolen'' ($s_4$) and ``absent ($s_5$) signals the generator to update the scene configuration, ensuring long-range logical consistency.

\begin{algorithm}[t]
\small
\caption{CANVAS}
\label{alg:canvas_short}

\KwIn{Shots $S=\{s_1,\dots,s_T\}$}
\KwOut{Frames $I=\{I_1,\dots,I_T\}$}

\textbf{1. Global Plan:}
Extract characters, locations, objects $\rightarrow$ define states $(\mathcal{C},\mathcal{L},\mathcal{O})$.
Initialize memory $\mathcal{M}$ (characters, locations, frames).

\textbf{2. Sequential Generation:}
\For{$t=1$ to $T$}{
Retrieve required states from plan.

Construct anchor set $\mathcal{A}_t$ from:
character anchors, previous frame, or location memory.

Generate candidates $\{I_t^i\}$.

Score each using:
prompt alignment + character + background + prop consistency.

Select $I_t^* = \arg\max_i Score(I_t^i)$.

Update memory $\mathcal{M}$ with $I_t^*$.
}

\Return $I=\{I_1^*,\dots,I_T^*\}$
\label{algo:canvas}
\end{algorithm}

\subsection{Sequential Generation with Memory}
\label{subsec:generation}
While the global plan specifies how the story evolves, maintaining visual consistency requires reusing previously generated visual information. $\canvas$ therefore generates shots sequentially while maintaining a persistent world memory.

\paragraph{Visual State Memory.}
The system maintains a persistent visual memory denoted by $\mathcal{M}=(\mathcal{M}_c,\mathcal{M}_l,\mathcal{M}_f)$. Here $\mathcal{M}_c$ stores character appearance anchor images, $\mathcal{M}_l$ stores background anchor images associated with locations, and $\mathcal{M}_f$ stores previously generated frame images.

\paragraph{Anchor Retrieval.}
For each shot $s_t$, the system first queries the global plan $\mathcal{P}$ to determine the required character appearances $\mathcal{C}(c,t)$, the location identity $\mathcal{L}(s_t)$, and the object states $\mathcal{O}(o,t)$ (Algorithm~\ref{alg:anchor_retrieval}). 
Based on this information and the current memory $\mathcal{M}$, the system retrieves a set of conditioning anchors denoted by $\mathcal{A}_t$. 

If the shot directly continues the previous scene without structural changes, the system retrieves the previous frame together with the required character anchors using $A_t=\{I_{t-1}\}\cup\{\mathcal{M}_c\}$. 
This preserves spatial layout, camera orientation, and character placement across consecutive shots .

If the shot revisits an existing location but introduces a different camera viewpoint, the system retrieves the stored location anchor together with the required character anchors using $A_t=\{\mathcal{M}_l(\mathcal{L}(s_t))\}\cup\{\mathcal{M}_c(c,\mathcal{C}(c,t))\}$.

If the location appears for the first time and no background anchor exists in memory, generation proceeds using only the relevant character anchors.

\paragraph{Candidate Generation and Selection.}
Given the retrieved anchors $\mathcal{A}_t$ and the shot description $s_t$, the image generator produces a set of candidate frames $\{I_t^1,I_t^2,\ldots,I_t^N\}$ (Prompt in Table~\ref{tab:candidate_generation_prompt}). To select the most consistent frame, $\canvas$ uses a \textit{QA-based selector} that evaluates each candidate through a set of targeted visual questions derived from the global continuity plan (Algorithm~\ref{alg:qa_selection}).
These questions verify whether character appearances match $\mathcal{C}(c,t)$, object states follow $\mathcal{O}(o,t)$, and the environment structure corresponds to the location assignment $\mathcal{L}(s_t)$ and the shot-description alignment.
Each candidate is scored based on its answers to these consistency checks, and the final frame is selected as $I_t^*=\\argmax_i \text{Score}(I_t^i \mid s_t,\mathcal{P})$, 
where the scoring function aggregates QA-based evaluations to measure narrative alignment and consistency.

\paragraph{Memory Update.}
After selecting the final frame $I_t^*$, the system extracts updated anchors from the generated image and updates the memory before proceeding to the next shot (Algorithm~\ref{alg:memory_update}). If a character $c$ is clearly visible in the frame, the system extracts the appearance anchor and updates $\mathcal{M}_c(c,\mathcal{C}(c,t))$. If the visual structure of a location changes due to an object state transition specified by $\mathcal{O}(o,t)$, the system extracts a new background anchor and updates $\mathcal{M}_l(\mathcal{L}(s_t))$.
For objects that appear in the frame, the system also extracts visual anchors and updates the prop memory $\mathcal{M}_o(o,\mathcal{O}(o,t))$, storing the current visual state of the object. This allows later shots to retrieve the correct object appearance when the object reappears or when the scene must reflect a changed object state.
Finally, the selected frame $I_t^*$ is stored in $\mathcal{M}_f$ so that the next shot can reuse the frame when the scene continues.

The running example in Figure~\ref{fig:method} illustrates this process. In shot $s_1$, the museum gallery appears for the first time, so generation proceeds without a background anchor and the resulting frame $I_1^*$ initializes the gallery and character anchors in memory. In shot $s_2$, the system retrieves the previous frame and character anchors to preserve the gallery layout. In shot $s_3$, Lena enters a new location (the security room), so the scene is generated without a location anchor and the new environment is stored afterward. In shot $s_4$, Lena returns to the gallery wearing the security jacket; the system retrieves the gallery background and her updated appearance anchor. Because the artifact is stolen, the gallery background anchor is updated to reflect the empty display case. Finally, in shot $s_5$, the system retrieves the updated gallery anchor together with Lena's blue-dress appearance anchor.
Through this sequential cycle of retrieval, generation, and memory update, $\canvas$ preserves character identity, maintains consistent environments across recurring locations, and correctly reflects object state changes throughout the story.

\begin{table*}[t]
\centering
\tiny
\setlength{\tabcolsep}{4pt}

\begin{tabular}{l l | c c | c | c c c | c}
\toprule

Dataset & Method 
& \multicolumn{2}{c|}{Background Consistency}
& Props 
& \multicolumn{3}{c|}{Character} 
& Char Avg  \\

\cmidrule(lr){3-4}
\cmidrule(lr){6-8}

&
& Consecutive & Non-Consecutive
& PropCons
& FaceCons & ClothCons & BodyCons
& \\

\midrule

\multirow{5}{*}{ViStoryBench} 
& AutoStudio \citep{cheng2025autostudiocraftingconsistentsubjects}   & 3.42 & 3.68 & 3.51 & 3.70 & 3.50 & 3.80 & 3.67 \\
\citep{zhuang2025vistorybenchcomprehensivebenchmarksuite} & Story-Iter \citep{mao2026storyitertrainingfreeiterativeparadigm}  & 3.66 & 3.85 & 3.60 & 3.80 & 3.60 & 3.90 & 3.77 \\
& Story2Board \citep{dinkevich2025story2boardtrainingfreeapproachexpressive} & 3.20 & 3.43 & 3.18 & 3.60 & 3.40 & 3.60 & 3.53 \\
& Gemini-CT  \citep{comanici2025gemini25pushingfrontier}  & 4.52 & 4.00 & 4.56 & 4.05 & 4.12 & 3.98 & 4.05 \\
\rowcolor{green!12}
& \textbf{CANVAS (Ours)} & \textbf{4.83} & \textbf{4.83} & \textbf{4.88} & \textbf{4.25} & \textbf{4.52} & \textbf{4.12} & \textbf{4.30} \\

\midrule

\multirow{5}{*}{ST-Bench} 
& AutoStudio \citep{cheng2025autostudiocraftingconsistentsubjects}  & 3.33 & 3.58 & 3.38 & 3.60 & 3.40 & 3.70 & 3.57 \\
\citep{zhang2025storymemmultishotlongvideo} & Story-Iter \citep{mao2026storyitertrainingfreeiterativeparadigm}   & 3.70 & 3.85 & 3.78 & 3.80 & 3.60 & 3.90 & 3.77 \\
& Story2Board \citep{dinkevich2025story2boardtrainingfreeapproachexpressive} & 3.22 & 3.50 & 2.89 & 3.70 & 3.50 & 3.80 & 3.67 \\
& Gemini-CT \citep{comanici2025gemini25pushingfrontier}   & 4.62 & 3.99 & 4.41 & 4.05 & 3.92 & 3.98 & 3.98 \\
\rowcolor{green!12}
& \textbf{CANVAS (Ours)} & \textbf{4.94} & \textbf{4.73} & \textbf{4.73} & \textbf{4.18} & \textbf{4.45} & \textbf{4.08} & \textbf{4.24} \\

\midrule

\multirow{5}{*}{HardContinuity}
& AutoStudio \citep{cheng2025autostudiocraftingconsistentsubjects}  & 3.05 & 3.18 & 2.90 & 3.90 & 3.05 & 3.30 & 3.42 \\
& Story-Iter \citep{mao2026storyitertrainingfreeiterativeparadigm}  & 3.65 & 3.65 & 2.56 & 4.78 & 3.81 & 4.27 & 4.29 \\
& Story2Board  \citep{dinkevich2025story2boardtrainingfreeapproachexpressive} & 3.33 & 3.33 & 2.23 & 4.39 & 3.78 & 4.02 & 4.06 \\
& Gemini-CT \citep{comanici2025gemini25pushingfrontier}   & 4.48 & 3.83 & 4.06 & 4.63 & 4.27 & 4.27 & 4.39 \\
\rowcolor{green!12}
& \textbf{CANVAS (Ours)} & \textbf{4.88} & \textbf{4.88} & \textbf{4.19} & \textbf{4.95} & \textbf{4.91} & \textbf{4.87} & \textbf{4.91} \\

\bottomrule
\end{tabular}
\caption{
Comparison of storyboard generation systems under the \textbf{ContinuityEval} framework using gemini-2.5-Flash as Judge.
Background consistency is reported separately for consecutive shot transitions and non-consecutive scene reappearances.
PropCons measures identity consistency of movable objects.
Character continuity evaluates identity preservation using facial appearance, clothing attributes, and hair/body cues.
The final column reports the \textbf{average character consistency}. To avoid the risk of circularity/bias in generation and evaluation, we also compare methods using GPT5 as Judge and report results in Table~\ref{tab:continuityevalwithgtp}.
}
\label{tab:autoraterjudgement}
\end{table*}

\section{Experimental Setup}
\label{sec:experimentalsetup}
\paragraph{Datasets.} 
We evaluate \textsc{CANVAS} on two existing storyboard generation benchmarks: \textbf{ViStoryBench-Lite} \citep{zhuang2025vistorybenchcomprehensivebenchmarksuite} (20 stories) and \textbf{ST-Bench}, derived from StoryMem \citep{zhang2025storymemmultishotlongvideo} (30 stories). However, these datasets provide limited challenges
: recurring locations rarely reappear after long gaps, character appearance changes are uncommon, and spatial dependencies across future events are seldom required.
To better evaluate long-range continuity reasoning, we introduce \textbf{HardContinuityBench}, a dataset generated through GPT-5.2 prompting to create storyboard narratives with deliberately challenging continuity conditions. The dataset emphasizes recurring locations, long-range scene reappearances, character appearance changes, and occlusion across cinematic shot transitions (e.g., wide shots followed by close-ups), requiring models to preserve spatial layouts and character identity even when elements temporarily disappear. HardContinuityBench, though small, contains structured narratives and is intended to stress-test world-state consistency across extended shot sequences (Details in Appendix in Table~\ref{tab:hardbench_stats}).

\paragraph{Baselines and Implementation.}
We implement \textsc{CANVAS} using \textbf{Gemini-3-pro-image} as the backbone image generation model. 
Since $\canvas$ is \textit{training-free}, we compare it against recent state-of-the-art methods that operate under a similar inference-time setting from Table~\ref{tab:method_comparison}. 
Specifically, we select all training-free baselines from Table~\ref{tab:method_comparison} and, for fair comparison, re-implement \textbf{AutoStudio} \citep{cheng2025autostudiocraftingconsistentsubjects}, Story-Iter \citep{mao2026storyitertrainingfreeiterativeparadigm}, and Story2Board \citep{dinkevich2025story2boardtrainingfreeapproachexpressive} using the same Gemini backbone.
Apart from this, we also consider \textbf{Gemini-Contextual (Gemini-CT)} as a baseline where we prompt the model to generate the image per shot with the previous keyframe and character anchor image as the context. We choose this baseline since it achieves the best prompt-following performance on ViStoryBench-Lite  \citep{zhuang2025vistorybenchcomprehensivebenchmarksuite}.

All methods take the \textit{shot-level textual descriptions} of a story as input, along with \textit{canonical reference images} for each character when available. For datasets such as \textbf{ViStoryBench-Lite} \citep{zhuang2025vistorybenchcomprehensivebenchmarksuite}, these canonical character images are provided as part of the benchmark. However, for \textbf{ST-Bench} \citep{zhang2025storymemmultishotlongvideo} and our proposed \textbf{HardContinuityBench}, such references are not available. In these cases, we first generate canonical character images from their textual descriptions and use them as identity anchors to guide the generation of continuous keyframes across shots.

\paragraph{Automatic Evaluation Metrics.}

We evaluate \textsc{$\canvas$} using two complementary settings targeting different aspects of storyboard quality.
First, we introduce \textbf{ContinuityEval}, a fine-grained continuity evaluator that measures whether generated shots maintain a coherent visual world across camera cuts. Storyboard generation models often introduce inconsistencies such as character appearance drift, environment layout changes, or objects appearing/disappearing across shots. ContinuityEval is model-agnostic and shows strong agreement with human judgments ($0.74$ Fleiss’ $\kappa$). We select the two VLM-as-Judge evaluators with the highest Pearson correlation with human ratings (details in Appendix~\ref{app:continuityvalidate}, Table~\ref{tab:autorater_validation}). Unlike prior embedding-based metrics, ContinuityEval evaluates three dimensions: \textbf{character consistency} (face, clothing, body identity), \textbf{background consistency} (geometric stability of environments), and \textbf{prop consistency} (identity and placement of movable objects across shots) (Appendix in Table~\ref{tab:faceprompt}, Table~\ref{tab:clothingprompt}, Table~\ref{tab:hairprompt}, Table~\ref{tab:bg_prompt}, Table~\ref{tab:prop_prompt}).
Second, we perform a \textbf{benchmark-based evaluation} on ViStoryBench \citep{zhuang2025vistorybenchcomprehensivebenchmarksuite} to assess prompt alignment and frame quality using the benchmark’s standard metrics, enabling direct comparison with prior methods (Table~\ref{tab:vistorybench_results}).

\paragraph{Human Preference User Study.}
In addition to VLM-based automatic scoring, we conduct a human preference study to evaluate narrative consistency. We randomly sample 20 stories (60 scenes) from the benchmark datasets and construct pairwise comparisons between the stories generated by $\canvas$ and four baselines (Gemini, Story2Board, StoryIter, AutoStudio), resulting in 240 comparison pairs. Fifteen annotators evaluate randomly sampled pairs across four dimensions: \textbf{background consistency}, \textbf{prop consistency}, \textbf{character consistency}, and \textbf{overall preference} (Human Annotation Instructions in Appendix~\ref{app:human_eval}).

\begin{figure}[!t]
    \centering
    \includegraphics[width=\linewidth]{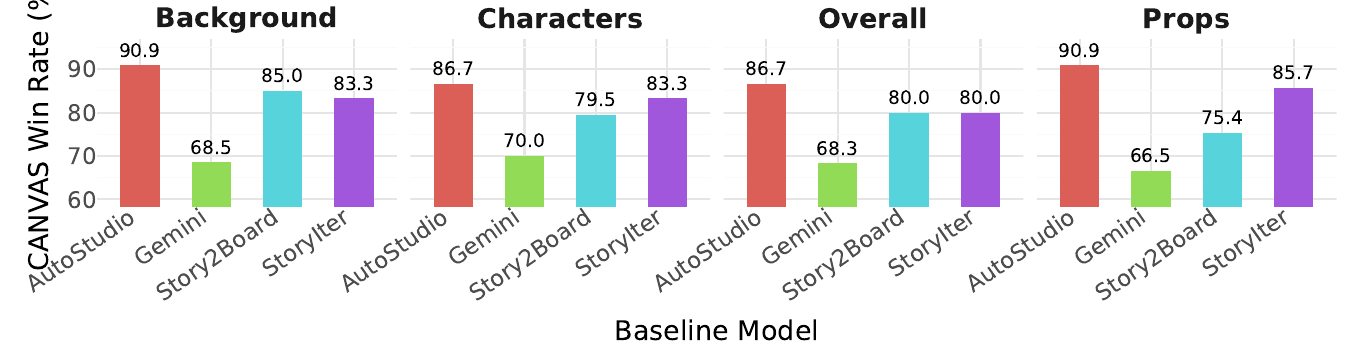}
    \caption{
Human pairwise preference win-rates of $\canvas$ against baseline storyboard generation systems across four continuity dimensions: background, props, characters, and overall coherence. Win-rates are computed on 240 side-by-side comparison of scenes and aggregated across 15 annotators. }
    \label{fig:humaneval}
\end{figure}

\begin{table}[t]
\centering
\scriptsize
\setlength{\tabcolsep}{4pt}
\begin{tabular}{lcccccccc}
\toprule

Method 
& \multicolumn{2}{c}{CSD $\uparrow$}
& \multicolumn{2}{c}{CIDS $\uparrow$}
& PA $\uparrow$
& CM $\uparrow$
& Inc $\uparrow$
& Aes $\uparrow$ \\

\cmidrule(lr){2-3}
\cmidrule(lr){4-5}

& Cross & Self 
& Cross & Self
& Avg
& & & \\

\midrule

Story-Iter
& 0.48 & 0.67 & 0.60 & 0.71 & 3.39 & 57.34 & 14.33 & 4.55 \\

Story2Board
& 0.23 & 0.45 & 0.35 & 0.39 & 2.92 & 44.44 & 11.93 & 4.56 \\

AutoStudio
& 0.34 & 0.63 & 0.44 & 0.46 & 3.39 & 56.22 & 14.34 & 3.45 \\

Gemini-CT
& 0.38 & 0.62 & 0.58 & 0.65 & 3.61 & 59.97 & 12.50 & 5.54 \\

\midrule

\textbf{$\canvas$}
& \textbf{0.49} & \textbf{0.72}
& \textbf{0.61} & \textbf{0.71}
& 3.60
& \textbf{60.34}
& \textbf{13.89}
& \textbf{5.60} \\

\bottomrule
\end{tabular}

\caption{
Comparison on ViStoryBench-Lite. Metrics include Style Diversity (CSD), 
Character Identity Similarity (CIDS), Prompt Alignment (PA), OCCM (CM), 
Inception (Inc), and Aesthetic score (Aes).
}

\label{tab:vistorybench_results}

\end{table}

\begin{table}[t]
\centering
\setlength{\tabcolsep}{5pt}
\begin{tabular}{lcccc}
\toprule
Method & FaceCons & ClothCons & BodyCons & Char Avg \\
\midrule
$\canvas$ (Full) & \textbf{4.18} & \textbf{4.45} & \textbf{4.08} & \textbf{4.24} \\
-- Canonical Character Anchor & 2.55 & 2.76 & 2.47 & 2.59 \\
-- Character Anchor Retrieval & 4.10 & 4.08 & 3.96 & 4.05 \\
-- Character Memory Update & 4.09 & 4.12 & 3.89 & 4.03 \\
\bottomrule
\end{tabular}
\caption{Character-memory ablation for $\canvas$. Removing canonical anchors causes the largest degradation in identity consistency, while disabling retrieval or memory updates leads to moderate drops.}
\label{tab:char_ablation}
\end{table}

\begin{table}[t]
\centering
\setlength{\tabcolsep}{5pt}
\begin{tabular}{lccc}
\toprule
Method & Consecutive & Non-Consecutive & Prop Cons \\
\midrule
$\canvas$ (Full) & \textbf{4.94} & \textbf{4.73} & \textbf{4.73} \\
-- Location Grouping & 2.45 & 3.34 & 4.45 \\
-- Future Planning & 4.76 & 4.11 & 3.55 \\
-- Background Reuse & 2.12 & 2.44 & 4.54 \\
-- Prop State Update & 4.34 & 4.33 & 3.76 \\
\bottomrule
\end{tabular}
\caption{Ablation analysis of $\canvas$ backgroud and prop-state memory components. Removing location grouping and background reuse severely degrades scene consistency across shots, while disabling prop planning or state updates reduces object state consistency in the narrative.}
\label{tab:memory_ablation}
\end{table}

\section{Main Results}
\label{sec:mainresults}
We evaluate $\canvas$ on multiple storyboard generation benchmarks to measure its ability to maintain long-range visual consistency across shots. Across all datasets and evaluation protocols, $\canvas$ consistently improves background stability, prop-state consistency, and character identity preservation compared to existing baselines.

\paragraph{Evaluation on ContinuityEval.}
Table~\ref{tab:autoraterjudgement} reports results using our ContinuityEval framework. Across all datasets, $\canvas$ consistently outperforms the strongest baseline (Gemini-CT) in background and prop consistency while also improving character identity.
On ViStoryBench-Lite \citep{zhuang2025vistorybenchcomprehensivebenchmarksuite} and ST-Bench \citep{zhang2025storymemmultishotlongvideo}, $\canvas$ improves consecutive background consistency by about 6--7\% and non-consecutive scene reappearances by about 12\%, while prop and character consistency improve by 6--7\%. On HardContinuityBench, which stresses long-range narrative reasoning, improvements are larger, reaching +8.9\% for consecutive backgrounds, +14.0\% for long-range scene reappearances, and +11.8\% for character consistency. 
Overall, $\canvas$ achieves 6--9\% gains in consecutive and 12--14\% improvements in non-consecutive scene consistency.

\paragraph{Evaluation on Standard ViStoryBench Metrics.}
Table~\ref{tab:vistorybench_results} compares $\canvas$ with recent training-free storyboard generation methods on ViStoryBench-Lite \citep{zhuang2025vistorybenchcomprehensivebenchmarksuite}. $\canvas$ achieves the best overall performance across most metrics, improving style consistency and character consistency while maintaining strong prompt alignment and visual quality. These results indicate that the proposed continuity planning and memory mechanisms effectively preserve narrative coherence without sacrificing visual fidelity.

\paragraph{User Study Results.}
Figure~\ref{fig:humaneval} reports human pairwise preference win-rates of $\canvas$ against baseline methods across four dimensions: background continuity, prop consistency, character identity, and overall coherence. Across all dimensions, $\canvas$ is consistently preferred by annotators, achieving win-rates up to \textbf{90.9\%} for background and prop consistency, \textbf{86.7\%} for character consistency, and \textbf{86.7\%} overall when compared against AutoStudio, and \textbf{68.3\%} against our strongest baseline.
These results indicate that human evaluators strongly prefer $\canvas$-generated storyboards on all criteria.

\begin{figure}[!t]
    \centering
    \includegraphics[width=\linewidth]{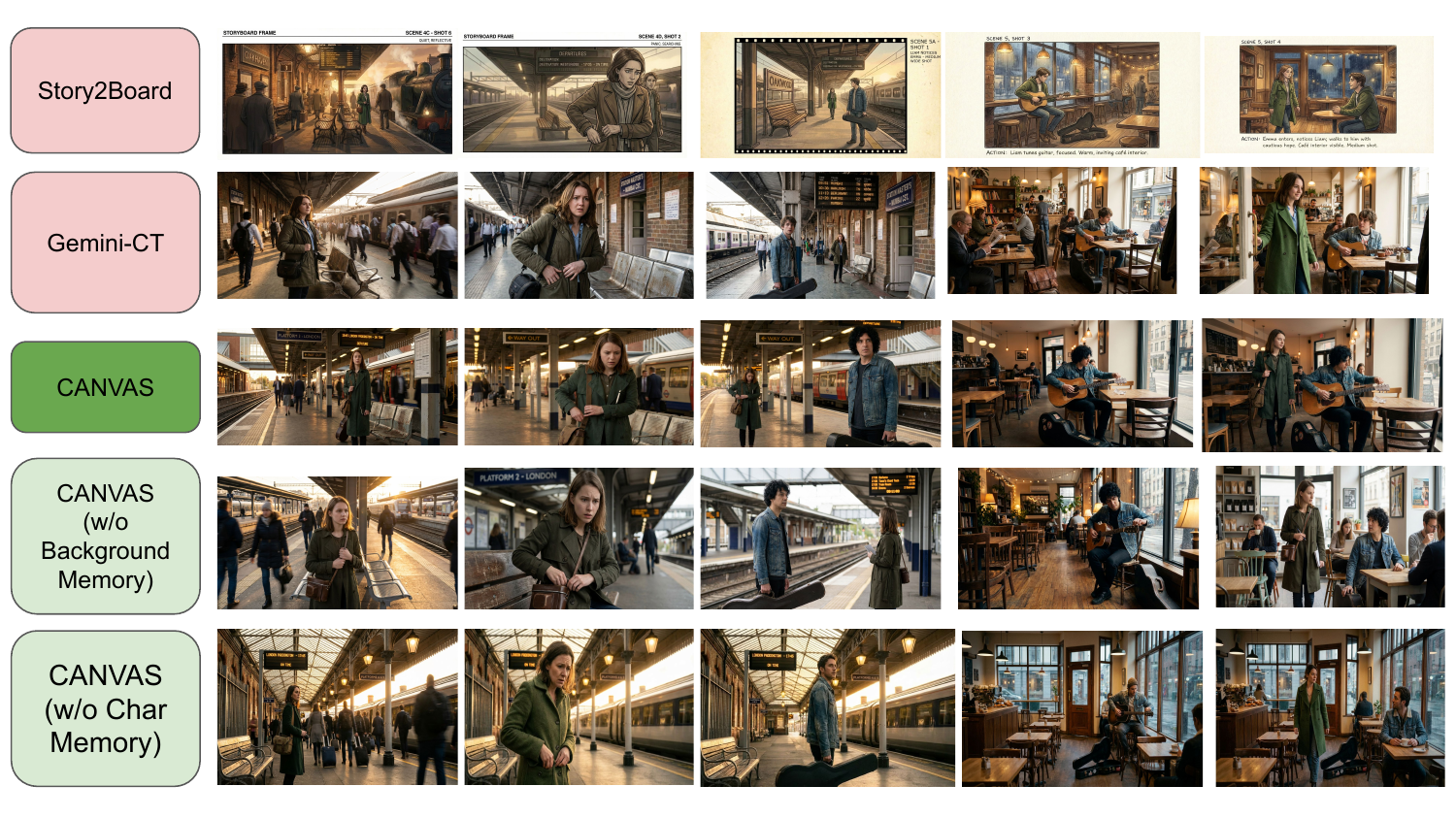}
    \caption{
Qualitative Examples of Train Story from HardContinuityBench along with ablation showing the importance of Background and Character Memory Update in CANVAS.}
\label{fig:trainexample}
\end{figure}

\section{Further Analysis}
\label{sec:ablation}

We ablate three main continuity mechanisms in \textbf{$\canvas$}: \textit{character memory}, \textit{background memory}, and \textit{prop-state memory}, on 20 sampled stories (including HardContinuityBench). Each variant removes one module while keeping the rest of the pipeline unchanged and is evaluated using \textbf{ContinuityEval}. Besides, we also ablate by removing QA selection techniques (Table~\ref{tab:autoraterjudgement_with_noqa}).

\paragraph{Character Memory.}
 Removing the \textbf{canonical character anchors} causes the largest degradation across all character consistency metrics, since the generator no longer has a fixed visual reference to preserve identity cues such as facial structure, clothing style, and body appearance. As a result, \textit{Face}, \textit{Cloth}, and \textit{Body} similarity drop by \textbf{39.0\%} (4.18$\rightarrow$2.55), \textbf{38.0\%} (4.45$\rightarrow$2.76), and \textbf{39.5\%} (4.08$\rightarrow$2.47), respectively. In contrast, disabling \textbf{anchor retrieval} leads to smaller declines—\textbf{1.9\%} (Face), \textbf{8.3\%} (Cloth), and \textbf{2.9\%} (Body)—since anchors exist but are not used during generation, while removing \textbf{memory updates} results in moderate drops of \textbf{2.2\%}, \textbf{7.4\%}, and \textbf{4.7\%}.

\paragraph{Background and Prop Memory.}
Background memory is critical for scene continuity: removing \textit{location grouping} or \textit{background reuse} reduces background consistency by up to \textbf{50--57\%}. Removing future planning mainly affects long-range scene and prop consistency, while disabling prop-state updates reduces object consistency by about \textbf{20\%}. Overall, these results confirm that $\canvas$’s memory mechanisms are essential for maintaining coherent environments and object states.

\paragraph{Effect of Candidate Variants.}
Increasing the number of generated variants mainly improves character consistency, as the selection step can choose candidates best aligned with character anchors. Background and prop consistency remain largely stable due to memory constraints (Figure~\ref{fig:propablationexample}). Performance saturates at $N=3$, providing the best balance between quality and computation.

\section{Qualitative Analysis}
Across the train-station storyline (Fig.~\ref{fig:trainexample}), prior approaches fail to preserve recurring environments and stable character identity. Story2Board alters station layout and signage across shots, $\canvas$ maintains consistent platform geometry, signage, and lighting while preserving the protagonist’s appearance.
In the laboratory storyline (Fig.~\ref{fig:labexample}), AutoStudio and Story-Iter fail to maintain object and environment persistence, with robot design, lab equipment, and corridor layouts changing across shots. $\canvas$ preserves consistency.

\paragraph{Ablation.} Removing background memory/planning causes location drift across shots, while removing character memory leads to identity changes in facial features and clothing (Fig.~\ref{fig:trainexample}). Removing prop updates breaks object-state continuity in the museum heist example (Fig.~\ref{fig:propablationexample}).

\section{Conclusion and Future Work}
In this paper, we introduce CANVAS, a training-free framework for generating coherent multi-shot visual storyboards from textual narratives by explicitly models long-range narrative continuity. 
The framework maintains three key dimensions of narrative coherence: character identity consistency, background continuity, and prop-state persistence. 
To systematically evaluate these properties, we also introduced HardContinuityBench and a fine-grained ContinuityEval framework that measures visual consistency across multiple dimensions of characters, environments, and objects.
In future, we plan to explore interactive human-in-the-loop visual storytelling, where users iteratively refine scenes, characters or provide natural language feedback while maintaining global continuity. 

\section*{Limitations}
Despite promising results, CANVAS has a few limitations.
First, the framework currently relies on large image generation models, making generation computationally expensive when producing multiple candidate shots for QA-based selection.
Second, although CANVAS improves continuity across characters, backgrounds, and props, it does not yet fully model fine-grained physical interactions or complex motion dynamics across shots.
Third, the system assumes that narrative entities such as characters, locations, and objects can be reliably extracted from text, which may be challenging for ambiguous or highly abstract narratives. 
Finally, the dataset size of HardContinuityBench is small, though we have used to stress-test our model, we plan to augment with more samples.

\paragraph{Use of AI assistants.} The authors used AI tools (OpenAI's ChatGPT, Google Gemini, Anthropic's Claude) for coding assistance during data analysis and visualization, and as a writing assistant limited to paraphrasing for conciseness. All substantive content, analysis, and conclusions are the authors' own work.
\section*{Ethical Considerations}
The data collection did not involve any crowdsourcing platforms. The study protocol was reviewed and approved by the institutional Ethics Review Board following a detailed assessment of the data collection procedures.
\paragraph{Potential Risks and Harms.}
While improving long-range visual consistency enhances narrative coherence, it may also enable the creation of more realistic synthetic content, raising concerns about misuse for misinformation, deepfakes, and fabricated visual evidence.
\bibliography{main}

\appendix

\section{Appendix Organization}
\label{sec:appendix}
We organize the appendix section to provide additional details to reproduce our pipeline:
\begin{itemize}
\item ContinuityEval Implementation Details (Appendix~\ref{app:continuityeval})
\item Computational Efficiency Analysis in Appendix~\ref{app:computational}
    \item Detailed Related Work in Appendix~\ref{app:further_related_work}
    \item Human Evaluation (User Study Instructions) in Appendix~\ref{app:human_eval}
    \item Impact of StoryBoard Quality on VideoGeneration using FilMaster Continuity and Expressiveness Measures (Appendix~\ref{subsubsec:filmEval})
    \item CANVAS Prompts and Implementation Algorithms and Case Study (Appendix~\ref{app:canvas})
    
    \item HardContinuityBench Creation Prompt, additional statistics of the dataset
    
    \item ContinuityEval Validation (Correlation with Human Judgement and Inter-Model Judgement Agreement) in Appendix in Table~\ref{tab:autorater_validation}
    
    \item More Ablation Analysis (without QA) in Table~\ref{tab:autoraterjudgement_with_noqa}
    
\end{itemize}

\subsection{Why HardContinuityBench is Hard?}
\label{app:hardcontinuityBench_justification}
Table~\ref{tab:hardbench_stats} highlights the structural differences between HardContinuityBench and existing storyboard benchmarks. Compared to ViStoryBench-Lite and ST-Bench, HardContinuityBench increases the likelihood of long-range dependencies such as recurring locations and delayed character reappearances by introducing significantly larger temporal gaps between scene reappearances, often requiring models to reconstruct environments after multiple intervening shots such as close-ups where the background is not visible. The dataset also includes more frequent character appearance changes and object state transitions, forcing systems to track evolving world states across the narrative. These characteristics make HardContinuityBench a substantially more challenging benchmark for evaluating long-range visual continuity in storyboard generation.

\begin{figure*}[!t]
    \centering
    \includegraphics[width=\linewidth]{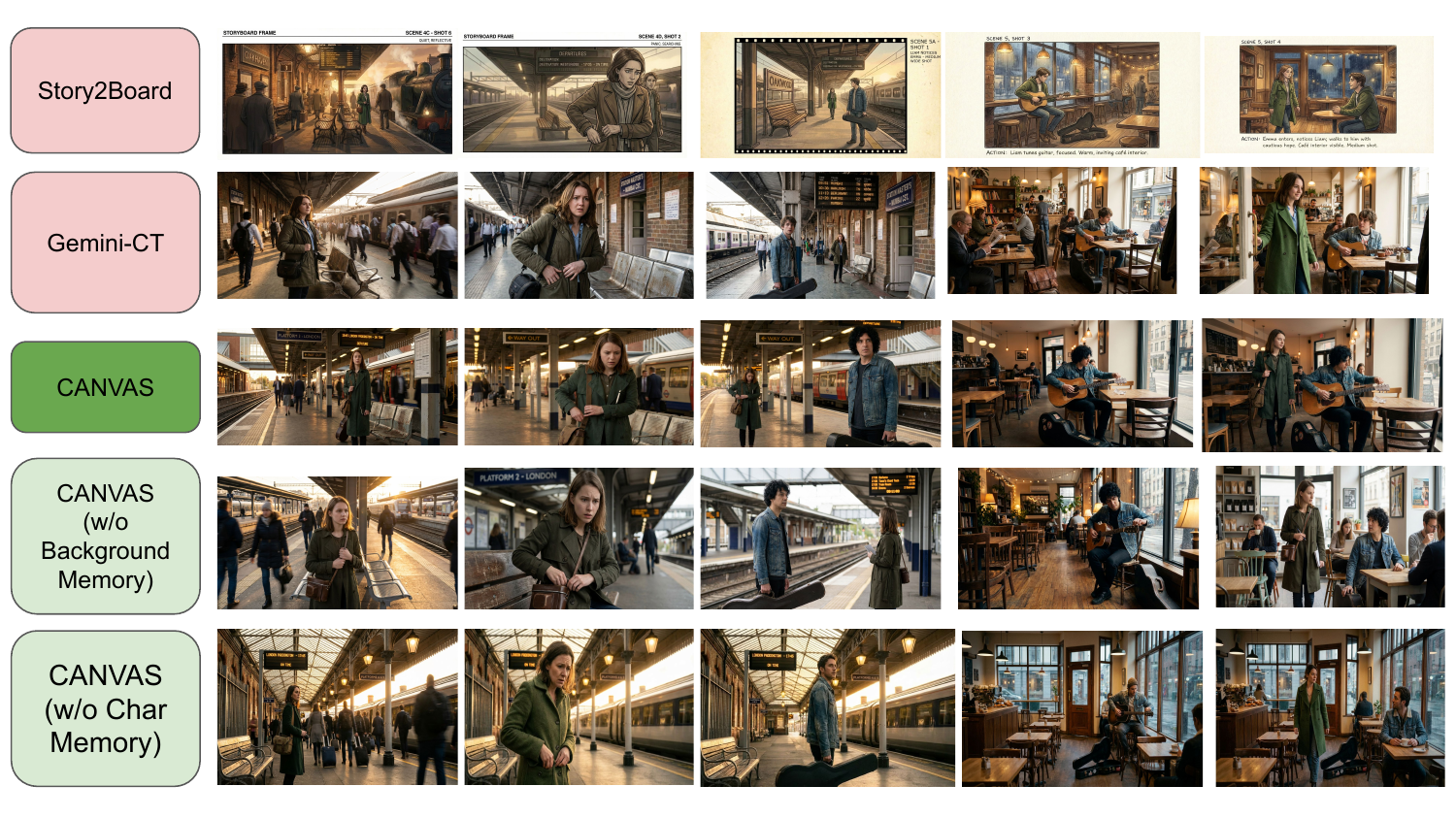}
    \caption{
Qualitative Examples of Train Story (Zoomed-in) from HardContinuityBench along with ablation showing the importance of Background and Character Memory Update in CANVAS.}
\label{fig:zoomedtrain}
\end{figure*}

\subsection{Character Consistency Evaluation}
\label{app:continuityeval}
To measure whether characters remain visually consistent across a generated storyboard, we design a Character Consistency Evaluator as part of the ContinuityEval framework. The goal of this evaluator is to quantify how well a character preserves identity attributes across multiple appearances throughout the narrative.

Given a sequence of generated storyboard frames $I_1, I_2, ..., I_T$ and corresponding shot descriptions $S_1, S_2, ..., S_T$, we first extract all characters mentioned in the shot descriptions using a language model. Character mentions referring to the same individual are canonicalized into a unified identity representation. For each character $c$, we then construct a temporal appearance sequence

\[
T_c = (I_{t_1}, I_{t_2}, ..., I_{t_k})
\]

where $I_{t_i}$ denotes the $i$-th frame in which character $c$ appears. The sequence represents the chronological timeline of that character's appearances across the storyboard.

Character continuity is evaluated by comparing consecutive appearances of the same character in the timeline unless appearance change is required. 
For each pair of adjacent frames $(I_{t_i}, I_{t_{i+1}})$, a Vision–Language Model evaluates identity consistency using three attribute groups: facial identity, clothing attributes, and hair/body appearance. Each attribute is evaluated independently using dedicated prompts and scored on a Likert scale from 1 to 5.

The facial identity evaluator compares structural facial features including facial shape, eye configuration, nose structure, jawline, facial proportions, and facial hair when visible. Differences caused by viewpoint, expression, lighting, or partial occlusion are ignored (Prompt in Table~\ref{tab:faceprompt}). The clothing evaluator measures consistency of garment category, clothing structure, layering, dominant color palette, and fabric appearance. Minor changes caused by lighting or folds are ignored (Prompt in Table~\ref{tab:clothingprompt}). The hair and body evaluator measures consistency of hairstyle, hair length, hair color, body silhouette, and wearable accessories such as glasses, hats, or backpacks (Prompt in Table~\ref{tab:hairprompt}).

\begin{table*}[t]
\centering
\setlength{\fboxsep}{6pt}

\begin{tabular}{|p{0.95\linewidth}|}
\hline
\textbf{Face Identity Evaluation Prompt.}

You are given two storyboard frames, Frame A and Frame B. The same character is expected to appear in both frames. Determine whether the faces correspond to the same individual.

Focus only on the specified character. Ignore other people and background objects.

Compare the following facial attributes: facial shape, eye shape and spacing, nose structure, jawline and chin structure, facial proportions, skin tone, and facial hair if present.

Ignore changes caused by camera angle, facial expression, lighting conditions, or partial occlusion.

Assign a score using the following rules:

5 — All visible facial features match. Facial structure, proportions, and distinctive features are consistent across both frames.

4 — All major facial features match. Minor variations appear only in rendering detail.

3 — At least half of the visible facial attributes match, but other attributes differ or are unclear.

2 — Fewer than half of the facial attributes match.

1 — Most visible facial attributes do not match.

If the face of the character is not visible in one or both frames, return \texttt{null}.
\\ \hline
\end{tabular}

\caption{Prompt used for Calculating Facial Identity Consistency.}
\label{tab:faceprompt}

\end{table*}

\begin{table*}[t]
\centering
\setlength{\fboxsep}{6pt}

\begin{tabular}{|p{0.95\linewidth}|}
\hline
\textbf{Clothing Consistency Evaluation Prompt.}

You are given two storyboard frames containing the same character. Evaluate whether the clothing worn by the character is consistent across the frames.

Focus only on the clothing worn by the character.

Compare the following attributes: garment category (coat, jacket, shirt, dress, uniform), clothing structure (sleeve length, coat length, collar type), clothing layers (for example jacket over shirt), dominant colors, patterns or logos, and fabric appearance.

Ignore differences caused only by lighting changes or cloth wrinkles.

Assign a score using the following rules:

5 — Garment type, structure, color palette, and patterns are identical.

4 — Garment type and structure are identical, and color palette differs only slightly.

3 — Garment type matches but structural details or colors differ.

2 — Garment type differs but some colors or patterns match.

1 — Garment type, structure, and colors are completely different.
\\ \hline
\end{tabular}

\caption{Prompt used for Calculating Clothing Similarity Consistency.}
\label{tab:clothingprompt}

\end{table*}

\begin{table*}[t]
\centering
\tiny
\setlength{\tabcolsep}{6pt}

\begin{tabular}{lccc}
\toprule
\textbf{Statistic} & \textbf{ViStoryBench-Lite} & \textbf{ST-Bench} & \textbf{HardContinuityBench} \\
\midrule

Stories & 20 & 30 & 5 \\

Recurring Location Reappearances & Rare & Occasional & \textbf{Frequent} \\

Avg. Gap Between Location Reappearances (shots) & 1.4 & 2.3 & \textbf{5.6} \\

Character Appearance Changes & Rare & Limited & \textbf{Common} \\

Prop State Transitions (per story) & 0.8 & 1.2 & \textbf{3.6} \\

Long-Range Scene Reconstruction Required & Low & Moderate & \textbf{High} \\

\bottomrule
\end{tabular}

\caption{Dataset statistics comparing HardContinuityBench with existing storyboard benchmarks. HardContinuityBench contains longer narratives, more recurring environments, larger temporal gaps between scene reappearances, and more object state transitions. These properties require models to maintain persistent world representations of characters, locations, and objects rather than relying only on short-range frame conditioning.}
\label{tab:hardbench_stats}
\end{table*}

\begin{table*}[t]
\centering
\setlength{\fboxsep}{6pt}
\tiny
\begin{tabular}{|p{0.95\linewidth}|}
\hline
\textbf{Background Architecture Consistency Evaluation Prompt.}

You are an expert evaluator for cinematic storyboard continuity.

You are given **two consecutive storyboard frames from the same scene in a narrative sequence**.

Your task is to evaluate **Background Geometry Consistency (GeomCons)** between the two frames.

GeomCons measures whether the **architectural structure of the environment remains logically coherent across shots**.

Your evaluation must focus **only on architectural elements** and determine whether the spatial layout of the environment remains consistent between the two frames.

---

 Important Instructions

Do **NOT** evaluate the following:

- characters
- character pose or movement
- facial expression
- lighting differences
- color shifts
- camera blur
- small objects or movable props
- decorations

These elements are **irrelevant to this evaluation**.

Your focus must be **only architectural elements that define the physical structure of the environment**.

---

Definition of Architectural Elements

Architectural elements are **fixed structural components of the environment** that define the geometry of the scene.

Examples include:

- walls
- doorways
- doors
- windows
- pillars
- columns
- arches
- staircases
- floors
- ceilings
- railings
- architectural partitions
- built-in counters
- permanent architectural installations

These elements define the **structural layout and geometry of the environment**.

Do **NOT** include movable objects such as chairs, books, tools, decorations, or equipment.

---

Step 1 — Identify Architectural Elements in Frame A

Inspect **Frame A** (the earlier frame).

List all architectural elements visible in the scene.

Examples:

- left wall
- right wall
- doorway
- window
- pillar
- staircase
- railing
- floor boundary
- ceiling beam

Let the number of architectural elements be:

---

 Step 2 — Identify Architectural Elements in Frame B

Inspect **Frame B** (the later frame).

List all architectural elements visible in the scene.

Let the number of elements be:

---

Step 3 — Match Corresponding Architectural Elements

Match architectural elements that appear in **both frames**.

Example:

| Frame A | Frame B |
|--------|--------|
| left wall | left wall |
| doorway | doorway |
| pillar | pillar |

Let the number of matched elements be:

---

 Step 4 — Detect Structural Violations

Evaluate the following violations.

---

 Appearance Violations

Architectural elements appearing in Frame B that were **not visible or implied in Frame A**.

Count the number of such cases:

---

Disappearance Violations

Architectural elements visible in Frame A that **disappear in Frame B** without plausible explanation.

Count the number of such cases:

---

 Layout Violations

Architectural elements that **shift position or orientation in physically impossible ways**.

Count the number of such cases:

---

Step 5 — Evaluate Camera Motion Plausibility

Camera movement may explain some changes.

Acceptable explanations include:

- camera pan
- camera zoom
- camera reframing
- camera moving closer
- camera moving sideways

If the change **cannot be explained by camera motion**, it should be counted as a violation.

---

Geometry Consistency Percentage

First compute the total number of violations:
Total Violations =
$N_appear$ + $N_disappear$ + $N_layout$

Normalize by the total number of architectural elements:

Max Elements = max($N_A$, $N_B$)

Compute the **Geometry Consistency Percentage**:

GeomCons = 100 × (1 minus Total Violations / Max Elements)

Clamp the result between **0 and 100**.

---
 Score Conversion (Percentage → Rubric)

Convert the percentage to a **GeomCons score from 1–5**.

| Percentage Range | Score | Interpretation |
|------------------|------|----------------|
| **90–100\%** | 5 | Perfect architectural consistency |
| **75–89\%** | 4 | Minor geometric differences |
| **55–74\%** | 3 | Moderate inconsistencies |
| **30–54\%** | 2 | Major structural inconsistencies |
| **0–29\%** | 1 | Severe spatial inconsistency |

\\ \hline
\end{tabular}

\caption{Prompt for Background Geometry Consistency Evaluation (In Consecutive and Non-Consecutive Shots).}
\label{tab:clothingprompt}

\end{table*}

\begin{table*}[t]
\centering
\small
\setlength{\tabcolsep}{5pt}

\begin{tabular}{p{3cm} p{11cm}}
\toprule

\textbf{Component} & \textbf{Description} \\

\midrule

Evaluation Goal &
Evaluate \textbf{Prop Consistency (PropCons)} between two storyboard frames by measuring how many props visible in one frame can be correctly recovered in the other frame with identical visual identity. \\

Evaluation Scope &
The evaluation must rely \textbf{only on visually observable information}. Objects that are not visible in one of the frames must not be considered. \\

Visible Prop Constraint &
If a prop is hidden due to camera angle, cropping, or occlusion, it must not be assumed to exist in the other frame. Only props that are clearly visible in both frames can be evaluated. \\

Matching Criteria &
Two props are considered identical only if the following attributes match:
shape, color, and texture/material appearance. \\

Ignored Factors &
The evaluation ignores character identity, character motion, lighting differences, motion blur, and architectural structures. \\

Definition of Props &
Props are \textbf{movable non-structural objects} present in the scene. Examples include chairs, tables, lamps, books, tools, decorations, statues, paintings, containers, and equipment. \\

Excluded Objects &
Structural elements such as walls, doors, windows, floors, pillars, and architectural structures must not be included in the prop evaluation. \\

Step 1 &
Identify all props clearly visible in Frame A. Let the number of visible props be $N_A$. \\

Step 2 &
Identify all props clearly visible in Frame B. Let the number of visible props be $N_B$. \\

Step 3 &
Determine which props can be correctly matched across both frames. A prop is considered recoverable only if its shape, color, and texture are identical in both frames. Let the number of matched props be $N_{match}$. \\

Step 4 &
Compute prop recoverability using the smaller visible prop set:

\[
ReferenceProps = \min(N_A, N_B)
\]

\[
Recoverability = 100 \times \frac{N_{match}}{ReferenceProps}
\]

Clamp the value between 0 and 100. \\

\bottomrule
\end{tabular}

\caption{Prop Consistency (PropCons) evaluation protocol used in ContinuityEval. The evaluator measures the recoverability of movable scene props between two frames based strictly on visible object identity.}
\label{tab:propcons_protocol}

\end{table*}

\begin{table*}[t]
\centering
\setlength{\fboxsep}{6pt}

\begin{tabular}{|p{0.95\linewidth}|}
\hline
\textbf{Hair and Body Appearance Evaluation Prompt.}

You are given two storyboard frames containing the same character. Evaluate whether the character’s hairstyle and body-level appearance are consistent.

Compare the following attributes: hairstyle, hair length, hair color, hair texture, body silhouette or build, and wearable accessories such as glasses, hats, helmets, or backpacks.

Ignore differences caused by body pose, camera viewpoint, or partial occlusion.

Assign a score using the following rules:

5 — Hairstyle, hair color, body silhouette, and accessories are identical.

4 — Hairstyle and body silhouette are identical, with only minor accessory or lighting differences.

3 — Either hairstyle or body silhouette matches, but the other differs.

2 — Both hairstyle and body silhouette differ but one accessory matches.

1 — Hairstyle, body silhouette, and accessories are all different.
\\ \hline
\end{tabular}

\caption{Prompt used for Calculating HairStyle and Body Cues Consistency.}
\label{tab:hairprompt}

\end{table*}

For each frame transition $(I_{t_i}, I_{t_{i+1}})$ the evaluator produces three attribute scores:

\[
f_i = \text{FaceScore}(I_{t_i}, I_{t_{i+1}}, c)
\]

\[
cl_i = \text{ClothingScore}(I_{t_i}, I_{t_{i+1}}, c)
\]

\[
hb_i = \text{HairBodyScore}(I_{t_i}, I_{t_{i+1}}, c)
\]

If an attribute cannot be evaluated due to severe occlusion or cropping, the evaluator returns a null value and that attribute is excluded from aggregation. The transition identity score is then computed as the mean of the available attribute scores:

\[
score_i = \text{mean}(f_i, cl_i, hb_i)
\]

For a character appearing in $k$ frames, the overall character consistency score is computed as the average of the transition scores across the timeline:

\[
CIDS(c) =
\frac{1}{k-1}
\sum_{i=1}^{k-1} score_i
\]

where $k-1$ represents the number of temporal transitions for the character.

Finally, the overall Character Identity Consistency Score for the story is obtained by averaging across all characters appearing in the narrative:

\[
CharacterConsistency =
\frac{1}{|C|}
\sum_{c \in C} CIDS(c)
\]

where $C$ denotes the set of characters present in the storyboard.

The resulting score lies in the range $[1,5]$, where higher values indicate stronger visual identity preservation across shots and lower values indicate character appearance drift or identity inconsistency.

\begin{table*}[t]
\centering
\tiny
\setlength{\tabcolsep}{4pt}

\begin{tabular}{l l | c c | c | c c c | c}
\toprule

Dataset & Method 
& \multicolumn{2}{c|}{Background Consistency}
& Props 
& \multicolumn{3}{c|}{Character} 
& Char Avg  \\

\cmidrule(lr){3-4}
\cmidrule(lr){6-8}

&
& Consecutive & Non-Consecutive
& PropCons
& FaceCons & ClothCons & BodyCons
& \\

\midrule

\multirow{6}{*}{ViStoryBench} 
& AutoStudio \citep{cheng2025autostudiocraftingconsistentsubjects}   & 3.42 & 3.68 & 3.51 & 3.70 & 3.50 & 3.80 & 3.67 \\
\citep{zhuang2025vistorybenchcomprehensivebenchmarksuite} & Story-Iter \citep{mao2026storyitertrainingfreeiterativeparadigm}  & 3.66 & 3.85 & 3.60 & 3.80 & 3.60 & 3.90 & 3.77 \\
& Story2Board \citep{dinkevich2025story2boardtrainingfreeapproachexpressive} & 3.20 & 3.43 & 3.18 & 3.60 & 3.40 & 3.60 & 3.53 \\
& Gemini-CT  \citep{comanici2025gemini25pushingfrontier}  & 4.52 & 4.00 & 4.56 & 4.05 & 4.12 & 3.98 & 4.05 \\
\rowcolor{gray!12}
& CANVAS w/o QA & 4.79 & 4.78 & 4.84 & 4.12 & 4.30 & 4.01 & 4.14 \\
\rowcolor{green!12}
& \textbf{CANVAS (Ours)} & \textbf{4.83} & \textbf{4.83} & \textbf{4.88} & \textbf{4.25} & \textbf{4.52} & \textbf{4.12} & \textbf{4.30} \\

\midrule

\multirow{6}{*}{ST-Bench} 
& AutoStudio \citep{cheng2025autostudiocraftingconsistentsubjects}  & 3.33 & 3.58 & 3.38 & 3.60 & 3.40 & 3.70 & 3.57 \\
\citep{zhang2025storymemmultishotlongvideo} & Story-Iter \citep{mao2026storyitertrainingfreeiterativeparadigm}   & 3.70 & 3.85 & 3.78 & 3.80 & 3.60 & 3.90 & 3.77 \\
& Story2Board \citep{dinkevich2025story2boardtrainingfreeapproachexpressive} & 3.22 & 3.50 & 2.89 & 3.70 & 3.50 & 3.80 & 3.67 \\
& Gemini-CT \citep{comanici2025gemini25pushingfrontier}   & 4.62 & 3.99 & 4.41 & 4.05 & 3.92 & 3.98 & 3.98 \\
\rowcolor{gray!12}
& CANVAS w/o QA & 4.89 & 4.70 & 4.69 & 4.08 & 4.24 & 3.98 & 4.10 \\
\rowcolor{green!12}
& \textbf{CANVAS (Ours)} & \textbf{4.94} & \textbf{4.73} & \textbf{4.73} & \textbf{4.18} & \textbf{4.45} & \textbf{4.08} & \textbf{4.24} \\

\midrule

\multirow{6}{*}{HardContinuity}
& AutoStudio \citep{cheng2025autostudiocraftingconsistentsubjects}  & 3.05 & 3.18 & 2.90 & 3.90 & 3.05 & 3.30 & 3.42 \\
& Story-Iter \citep{mao2026storyitertrainingfreeiterativeparadigm}  & 3.65 & 3.65 & 2.56 & 4.78 & 3.81 & 4.27 & 4.29 \\
& Story2Board  \citep{dinkevich2025story2boardtrainingfreeapproachexpressive} & 3.33 & 3.33 & 2.23 & 4.39 & 3.78 & 4.02 & 4.06 \\
& Gemini-CT \citep{comanici2025gemini25pushingfrontier}   & 4.48 & 3.83 & 4.06 & 4.63 & 4.27 & 4.27 & 4.39 \\
\rowcolor{gray!12}
& CANVAS w/o QA & 4.84 & 4.83 & 4.16 & 4.72 & 4.61 & 4.55 & 4.63 \\
\rowcolor{green!12}
& \textbf{CANVAS (Ours)} & \textbf{4.88} & \textbf{4.88} & \textbf{4.19} & \textbf{4.95} & \textbf{4.91} & \textbf{4.87} & \textbf{4.91} \\

\bottomrule
\end{tabular}

\caption{
Comparison of storyboard generation systems under the \textbf{ContinuityEval} framework using gemini-2.5-Flash as Judge.
Background consistency is reported separately for consecutive shot transitions and non-consecutive scene reappearances.
PropCons measures identity consistency of movable objects.
Character continuity evaluates identity preservation using facial appearance, clothing attributes, and hair/body cues.
The final column reports the \textbf{average character consistency}. 
We additionally report \textbf{CANVAS w/o QA}, which removes the candidate verification module. 
Even without QA-based selection, CANVAS remains competitive with or stronger than prior baselines, indicating that the gains largely arise from the continuity planning and memory mechanisms.
}
\label{tab:autoraterjudgement_with_noqa}
\end{table*}

\begin{figure}[!t]
    \centering
    \includegraphics[width=0.7\linewidth]{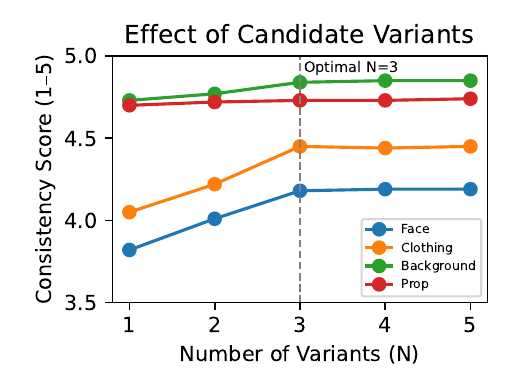}
    \caption{
Effect of generating N candidates on the ContinuityEval metrics.}
\end{figure}

\begin{figure*}[!t]
    \centering
    \includegraphics[width=\linewidth]{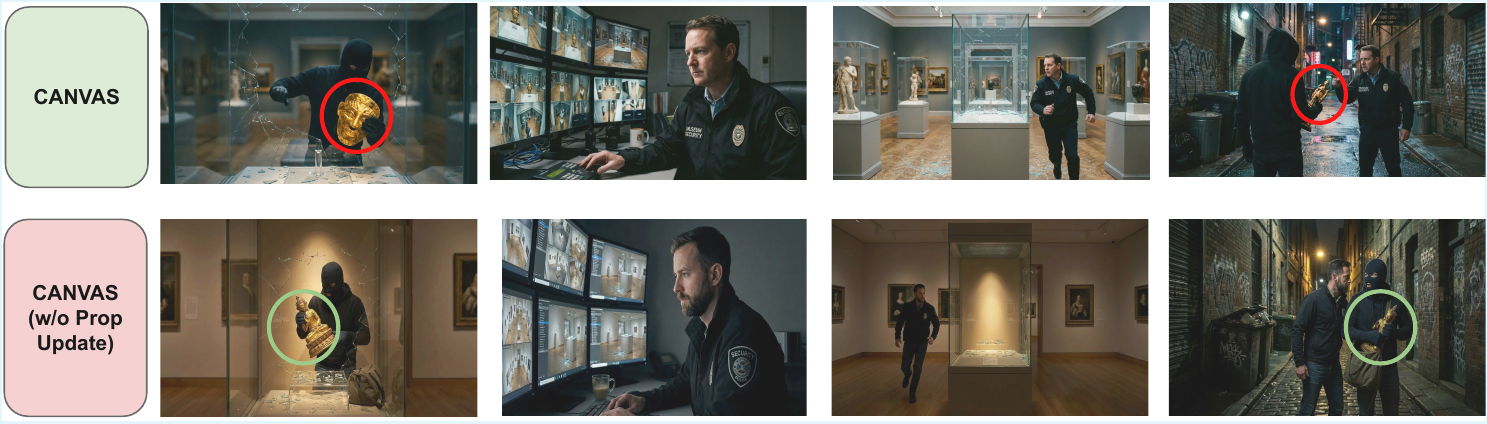}
    \caption{
Qualitative Ablation showing the importance of Prop-State Tracking in CANVAS.}
\label{fig:propablationexample}
\end{figure*}

\subsection{Computational Efficiency}
\label{app:computational}
All evaluated pipelines operate in a \textit{training-free inference setting} where the heavy computation (image generation and reasoning) is performed through hosted multimodal APIs. As a result, the local system only performs lightweight orchestration such as prompt construction, memory updates, and image I/O, allowing all methods to run on standard \textbf{CPU-only machines} without requiring local GPUs.

\paragraph{Story2Board.}
Story2Board generates one image per shot using a single API call, resulting in linear complexity $O(S)$ where $S$ is the number of shots. The method has minimal overhead since it performs no additional reasoning or verification.

\paragraph{StoryIter.}
StoryIter performs iterative refinement of frames across multiple passes. If $I$ refinement iterations are used, the pipeline requires $S(I+1)$ image-generation API calls. While the method includes local CLIP-based similarity computation for reference selection, this step is lightweight and can run on CPU.

\paragraph{CANVAS.}
CANVAS introduces additional reasoning and verification steps to enforce visual continuity. Each shot involves prop-state reasoning, multi-candidate image generation, and VLM-based candidate evaluation. With $K$ generated candidates per shot, the total number of API calls is approximately $1 + C + S(1+3K)$, where $C$ denotes the number of characters requiring canonical initialization. Although CANVAS incurs higher inference-time API usage, it maintains low local computational overhead and still runs efficiently on CPU-only systems.

Overall, the runtime of all pipelines is dominated by API latency rather than local computation, and all methods remain computationally lightweight compared to training-based approaches.

\subsection{Positioning of CANVAS}
\label{app:further_related_work}
\subsubsection{Visual Story Generation from Text.}
Early work on visual storytelling focused on generating sequences of images from textual narratives using generative models. One representative approach is StoryGen \citep{Liu_2024_CVPR}, which introduces an autoregressive diffusion-based framework for open-ended visual storytelling. The model generates each frame conditioned on the text description and preceding image-caption pairs using a vision–language context module integrated into Stable Diffusion.
During generation, features extracted from earlier frames are injected into the diffusion denoising process, allowing the model to maintain some degree of visual continuity across frames.

While StoryGen improves over independent text-to-image generation by conditioning on previous frames, its architecture primarily focuses on appearance-level consistency rather than explicit modeling of narrative state. Because generation is conditioned only on preceding frames, the system lacks a structured representation of characters, locations, or object states. Consequently, errors can accumulate over long sequences, and recurring environments or props are not guaranteed to remain stable across distant shots. In contrast, CANVAS introduces explicit world-state modeling with persistent character, background, and prop anchors, enabling reasoning about how entities evolve across the story.

Another line of work explores architectural modifications to diffusion models to improve cross-frame consistency. StoryDiffusion \citep{zhou2024storydiffusionconsistentselfattentionlongrange} introduces a consistent self-attention mechanism that injects reference tokens from previous frames into the attention computation of the diffusion model. This mechanism allows the model to maintain subject identity across generated frames while preserving the controllability of text prompts.
 However, StoryDiffusion \citep{zhou2024storydiffusionconsistentselfattentionlongrange}  focuses primarily on subject identity preservation, particularly character appearance and attire. The method does not attempt to model narrative structure, scene persistence, or object state evolution across the story. Moreover, because consistency is enforced through attention correlations rather than explicit memory structures, the model lacks a mechanism for reasoning about long-range narrative constraints, such as ensuring that objects change state only when story specifies such events.

More recent work explores training-free storyboard generation pipelines. Story2Board \citep{dinkevich2025story2boardtrainingfreeapproachexpressive} proposes a lightweight consistency mechanism combining latent panel anchoring and reciprocal attention value mixing to maintain subject identity across storyboard panels.
The system uses a language model to convert a story into panel-level prompts and then generates panels sequentially using diffusion models.

Although Story2Board \citep{dinkevich2025story2boardtrainingfreeapproachexpressive} improves panel-level coherence and allows dynamic pose and spatial variation, its design prioritizes visual expressiveness and layout diversity rather than strict narrative continuity. The framework does not maintain persistent representations of characters, locations, or props across panels, and therefore cannot explicitly enforce logical consistency when scenes reappear later in the story.

In contrast, CANVAS models story generation as an evolving narrative world state, enabling persistent tracking of characters, environments, and objects throughout the story timeline.

\subsubsection{Multi-Turn Image Generation and Interactive Story Creation}
Another research direction focuses on interactive image generation, where images are generated iteratively through multi-turn user instructions.

AutoStudio \citep{cheng2025autostudiocraftingconsistentsubjects} proposes a multi-agent framework for interactive image generation, combining large language models with diffusion models to maintain subject consistency across multiple turns. The architecture consists of a subject manager that interprets user instructions, a layout generator that predicts bounding boxes for subjects, a supervisor that refines layouts, and a diffusion-based image generator that produces the final image.
This design allows AutoStudio to maintain consistency for multiple subjects while responding to user instructions such as adding objects or modifying scene elements.
Despite these strengths, AutoStudio primarily addresses interactive editing and layout control rather than long-form narrative consistency. The system focuses on maintaining subject identity within short interactive sessions and does not explicitly model persistent environments or object states across a structured story. As a result, the framework does not address challenges such as maintaining stable environments when scenes reappear or ensuring that object states evolve logically according to narrative events.

Similarly, TheaterGen \citep{cheng2024theatergen} introduces a framework for multi-turn image generation in which a large language model acts as a ``screenwriter'' that generates a structured prompt book describing characters and layouts.
Character prompts are used to generate reference images during a rehearsal stage, and these references guide the final image generation during the diffusion process.
The key idea in TheaterGen is character management through prompt planning, which improves semantic consistency in multi-turn generation tasks.
However, TheaterGen primarily focuses on character-level consistency, and its architecture generates subjects individually before merging them into the final image. This design can lead to unnatural interactions between characters or inconsistencies in spatial relationships. Moreover, the framework does not maintain persistent representations of environments or objects across a story timeline.

Compared to these approaches, CANVAS focuses on story-level coherence rather than interactive editing, introducing explicit mechanisms to maintain background continuity, object persistence, and logical state transitions across narrative events.

\subsubsection{Memory-Based Narrative Generation}
Several recent works attempt to address long-range consistency using memory mechanisms.
StoryMem \citep{zhang2025storymemmultishotlongvideo} reformulates visual storytelling as iterative shot synthesis conditioned on a memory bank of keyframes from previously generated shots. The framework injects these keyframes into a single-shot video diffusion model using a Memory-to-Video architecture, enabling the model to preserve visual context across shots.
StoryMem improves cross-shot consistency by storing selected keyframes and using them as conditioning signals during generation.
However, the stored memory primarily functions as visual context for conditioning, rather than as an explicit representation of narrative entities. The system does not separately model characters, environments, and objects or reason about how these entities evolve across the story. As a result, the memory acts as a passive reference rather than an interpretable narrative representation.

Very recently, VideoMemory \citep{zhou2026videomemoryconsistentvideogeneration} introduces an entity-centric dynamic memory bank that stores representations of characters, props, and backgrounds. A multi-agent system retrieves these representations before generating each shot and updates the memory after generation.This architecture improves long-range entity consistency by maintaining persistent visual descriptors of story entities.
Nevertheless, VideoMemory primarily focuses on video-level entity persistence rather than storyboard-level reasoning about scene structure. The system retrieves stored representations but does not explicitly plan narrative continuity or enforce constraints such as maintaining stable environment geometry across recurring scenes. Furthermore, generation remains largely conditioned on visual descriptors rather than structured narrative reasoning.

Against Narrative Weaver \citep{yao2026narrativeweavercontrollablelongrange} and OneStory \citep{an2025onestorycoherentmultishotvideo}, HoloCine \citep{meng2025holocineholisticgenerationcinematic} which bake long-range context or planning into the generator, CANVAS’s advantage is its architecture-agnostic, training-free deployment. However, these training-based methods report strong long-range coherence and could serve as upper-bound references.

\begin{table*}[t]
\centering
\tiny
\setlength{\tabcolsep}{4pt}
\begin{tabular}{l|c|c|c|c|c|c|c}
\toprule
\textbf{Method} 
& \textbf{Generation Paradigm}
& \textbf{CharacterCons}
& \textbf{Environment}
& \textbf{Object State Tracking}
& \textbf{Interactive Editing}
& \textbf{Explicit Narrative}
& \textbf{Long-Range} \\

\midrule

StoryGen 
& Autoregressive diffusion 
& $\checkmark$
& $\xmark$
& $\xmark$
& $\xmark$
& $\xmark$
& $\xmark$ \\

StoryDiffusion 
& Attention-based diffusion consistency
& $\checkmark$
& $\xmark$
& $\xmark$
& $\xmark$
& $\xmark$
& $\xmark$ \\

AutoStory 
& Layout-guided story generation
& $\checkmark$
& $\triangle$
& $\xmark$
& $\xmark$
& $\xmark$
& $\xmark$ \\

Story2Board
& Training-free storyboard prompting
& $\checkmark$
& $\triangle$
& $\xmark$
& $\xmark$
& $\xmark$
& $\xmark$ \\

Story-Iter
& Iterative reference refinement
& $\checkmark$
& $\triangle$
& $\xmark$
& $\xmark$
& $\xmark$
& $\checkmark$ \\

TheaterGen
& LLM prompt-book + diffusion
& $\checkmark$
& $\xmark$
& $\xmark$
& $\checkmark$
& $\xmark$
& $\xmark$ \\

AutoStudio
& Multi-agent interactive generation
& $\checkmark$
& $\triangle$
& $\xmark$
& \checkmark
& $\xmark$
& $\triangle$ \\

StoryMem
& Keyframe memory conditioning
& $\checkmark$
& $\triangle$
& $\triangle$
& $\xmark$
& $\xmark$
& $\checkmark$ \\

VideoMemory
& Entity-centric memory bank
& $\checkmark$
& $\triangle$
& $\triangle$
& $\xmark$
& $\xmark$
& $\triangle$ \\

\midrule

\rowcolor{gray!10}
\textbf{CANVAS (Ours)}
& World-state narrative modeling
& $\checkmark$
& $\checkmark$
& $\checkmark$
& $\xmark$
& $\checkmark$
& $\checkmark$ \\

\bottomrule
\end{tabular}

\caption{
Positioning of \textbf{CANVAS} relative to prior visual storytelling and multi-turn image generation frameworks.
Most existing approaches enforce consistency through reference images or attention mechanisms, focusing primarily on character identity.
In contrast, \textbf{CANVAS} explicitly models the evolving \emph{narrative world state}, maintaining persistent representations of characters, environments, and object states across shots to enable long-range story continuity.
$\triangle$ indicates partial or implicit support.
}
\label{tab:method_positioning}
\end{table*}

\paragraph{Comparison Criteria.}
We compare methods along several dimensions capturing the key requirements of coherent visual storytelling in Table~\ref{tab:method_positioning}.

\textbf{Generation Paradigm.}
The primary architectural strategy used by the method to generate story images, such as autoregressive generation, attention-based consistency mechanisms, layout-guided generation, memory-conditioned generation, or explicit world-state modeling.

\textbf{Character Consistency.}
Whether the method explicitly preserves the visual identity of characters (e.g., face, clothing, body appearance) across multiple frames or shots.

\textbf{Environment Persistence.}
Whether the method ensures that recurring locations maintain stable spatial structure and visual layout when the scene reappears later in the story.

\textbf{Object State Tracking.}
Whether the system maintains and updates the state of narrative objects (e.g., a broken display case, a stolen artifact) across frames according to the story events.

\textbf{Interactive Editing.}
Whether the system supports multi-turn interaction where users iteratively modify previously generated images through dialogue or instructions.

\textbf{Explicit Narrative Modeling.}
Whether the framework maintains an explicit representation of the evolving narrative world (e.g., characters, environments, and object states) rather than relying only on visual similarity between generated frames.

\textbf{Long-Range Continuity.}
Whether the method can maintain consistency across distant frames or shots in long narratives where characters, environments, or objects reappear after several intermediate scenes.

\subsection{Human Evaluation Setup}
\label{sec:human_eval_setup}
While automatic metrics can measure certain aspects of visual similarity and prompt alignment, they often fail to capture more subtle narrative properties such as scene continuity, object persistence, and overall perceptual coherence across a storyboard. Therefore, we conduct a human evaluation study to assess whether generated storyboards preserve consistent environments, object states, and narrative structure from the perspective of human observers.

\paragraph{Annotators.}
We have asked \textbf{fifteen annotators}, consisting of members of our research group who have prior experience evaluating outputs of generative models. Participation in the study was voluntary.
Before the study begins, we provide them with detailed guidelines and example evaluations to ensure consistent interpretation of the evaluation criteria (Instructions in Section~\ref{app:human_eval}
and their interface is in Figure~\ref{fig:humaneval}). The annotators were blinded to the method identity and the order of sequences was randomized per trial.

\subsection{Human Annotation Instructions.}
\label{app:human_eval}
In this study, annotators compare two storyboard sequences generated by different methods (denoted as \textbf{Scene A} and \textbf{Scene B}) as shown in Figure~\ref{fig:humaneval}. Each storyboard contains a sequence of keyframes depicting the same narrative. The goal of the evaluation is to determine which method better preserves visual and narrative consistency. Please note that this is completely a voluntary help, and you can opt out if you do not want to help. All the data will be used for research purposes.

For each comparison, annotators are shown the full storyboard sequences for both methods. However, the evaluation focuses on a specific \textbf{target scene} (e.g., Scene$_4$). The remaining scenes are provided only as context for judging continuity.

Annotators must select one of the following options for each question:

\begin{itemize}
\item \textbf{Scene A}: Scene A performs better according to the criterion.
\item \textbf{Scene B}: Scene B performs better according to the criterion.
\item \textbf{Tie}: Both scenes perform equally well.
\end{itemize}

Annotators should evaluate only the target scene while using earlier scenes to assess consistency.

\paragraph{Background Consistency.}

This criterion evaluates whether the \textbf{environment or location} remains visually consistent with earlier scenes.
Annotators should check whether the scene maintains consistent spatial layout, stable environmental structure, persistent background elements (e.g., walls, windows, control panels, furniture), consistent lighting and overall scene configuration
If the story revisits a previously seen location, the environment should maintain the same overall layout and structure. Major changes such as disappearing walls, altered room geometry, or entirely different environments indicate poor background consistency.
Annotators should select the scene that better preserves the visual structure of the location across shots.

\paragraph{Prop Consistency.}

This criterion evaluates whether important \textbf{objects (props)} maintain the correct presence and state throughout the story.

Props include narrative-relevant objects such as tools, artifacts, equipment, documents, or other objects that appear repeatedly in the scene.

Annotators should verify that:

\begin{itemize}
\item objects that appear earlier remain present unless explicitly removed
\item objects that were removed or stolen do not reappear
\item objects maintain consistent visual appearance
\end{itemize}

Examples of inconsistencies include objects suddenly disappearing, reappearing after being removed, or changing form without narrative justification.

Annotators should select the scene that better preserves the correct object states.

\paragraph{Character Consistency.}

This criterion evaluates whether \textbf{character identity and appearance} remain consistent across scenes.

Annotators should examine whether characters maintain:
consistent facial identity, stable hairstyle and facial features, consistent clothing and accessories, consistent body appearance

Changes in appearance are acceptable only if they are explicitly described by the narrative (e.g., a character changing clothes). Otherwise, unexpected appearance changes should be considered inconsistencies.
Annotators should select the scene that better preserves character identity.

\paragraph{Overall Preference.}

This criterion measures the overall quality of the target scene, considering all aspects of visual storytelling.

Annotators should consider:

\begin{itemize}
\item alignment with the scene description
\item character identity preservation
\item background continuity
\item object state correctness
\item overall realism and coherence
\end{itemize}

Annotators should select the scene that provides the best overall visual depiction of the story.

\paragraph{Annotation Guidelines.}

When performing the evaluation, annotators should follow these guidelines:

\begin{itemize}
\item Evaluate only the \textbf{target scene} indicated in the interface.
\item Use the rest of the storyboard only to judge continuity.
\item Focus on major inconsistencies rather than minor stylistic differences.
\item Select \textbf{Tie} if both scenes are equally consistent.
\end{itemize}

\subsection{ContinuityEval Validation}
\label{app:continuityvalidate}
To verify that automated evaluations align with human judgments, we collect expert annotations from three evaluators experienced in image generation and visual evaluation. Annotators independently score storyboard frame pairs using the same rubrics as the autoraters, achieving an average inter-annotator agreement of \textbf{0.74 Fleiss' Kappa}. 

We then evaluate three multimodal models—\textbf{Gemini-2.5-Flash}, \textbf{Gemini-CT}, and \textbf{GPT-5}—using identical prompts. Pearson correlations between human and model scores are computed on 100 randomly sampled storyboard transitions from ViStoryBench and ST-Bench. As shown in Table~\ref{tab:autorater_validation}, automated evaluators exhibit strong agreement with human judgments across all continuity dimensions, with Gemini-2.5-Flash achieving the highest correlation.

Multi-shot storyboards require models to maintain a coherent visual world across camera cuts. However, generative models often introduce inconsistencies such as characters changing appearance, scene layouts shifting, or objects appearing and disappearing across shots. To obtain detailed and scalable measurements of narrative visual coherence, we design a \textbf{ContinuityEval-based evaluation framework} that provides fine-grained judgments of \textbf{character and background consistency} across generated storyboard sequences. The autorater evaluates each storyboard and assigns scores on a \textbf{1–5 Likert scale}, where 1 indicates severe inconsistency and 5 indicates perfect continuity.
ContinuityEval measures visual coherence through several structured dimensions.

\noindent
\textbf{[1] Character Consistency (\textit{CharCons}).}
Characters are the central entities of most narratives, and maintaining stable identity across shots is critical for visual storytelling. We therefore evaluate whether characters preserve consistent identity cues when they reappear in different frames. This dimension is decomposed into three components: \textbf{Facial Consistency (\textit{FaceCons})}, which measures whether facial structure and identity remain stable; \textbf{Clothing Consistency (\textit{ClothCons})}, which evaluates whether clothing attributes persist across shots; and \textbf{Body Appearance Consistency (\textit{BodyCons})}, which captures body-level cues such as hairstyle, body shape, and other global appearance attributes. 

\noindent
\textbf{[2] Non-Movable Object Consistency (\textit{GeomCons}).}
Even when characters remain consistent, storyboard frames may violate the physical structure of the environment—for example, walls, windows, or pillars may suddenly appear or disappear between shots. To capture such spatial errors, we evaluate whether the \textbf{non-movable architectural elements of the environment remain geometrically coherent across shots}. The prompt identifies structural elements such as walls, doors, windows, and pillars in both frames and detects violations including newly appearing structures, disappearing structures, or elements shifting to physically implausible positions without a plausible camera shift. A geometry consistency percentage is computed based on the number of such violations relative to the total number of architectural elements and is converted into a 1–5 score. 

\noindent
\textbf{[3] Movable Object Consistency (\textit{PropCons}).}
Storyboard continuity also requires maintaining the identity and placement of movable objects within the scene. Generative models often introduce inconsistencies where props suddenly appear, disappear, or change across shots. To measure this effect, we evaluate whether \textbf{movable objects (props) remain consistent across frames without sudden appearance or disappearance that cannot be explained by a plausible camera shift}. The prompt identifies visible props in both frames and determines which objects can be correctly matched based on identical shape, color, and texture. A prop recoverability percentage is computed based on how many props can be matched across frames, and this percentage is converted into a 1–5 PropCons score.
\begin{table}[!t]
\centering
\small
\setlength{\tabcolsep}{6pt}

\begin{tabular}{lccc}
\toprule
Model & Background & Character & Props \\
\midrule
Gemini-2.5-Flash & 0.71 & 0.68 & 0.67 \\
Gemini-3-Pro-Image    & 0.65 & 0.61 & 0.62 \\
GPT-5            & 0.70 & 0.66 & 0.68 \\
\bottomrule
\end{tabular}

\caption{Pearson correlation between human expert ratings and VLM-based autoraters across three evaluation dimensions. Scores are computed on 50 randomly sampled storyboard transitions from ViStoryBench and ST-Bench. }
\label{tab:autorater_validation}
\end{table}

\begin{figure*}[!t]
    \centering
    \fbox{\includegraphics[width=0.9\linewidth]{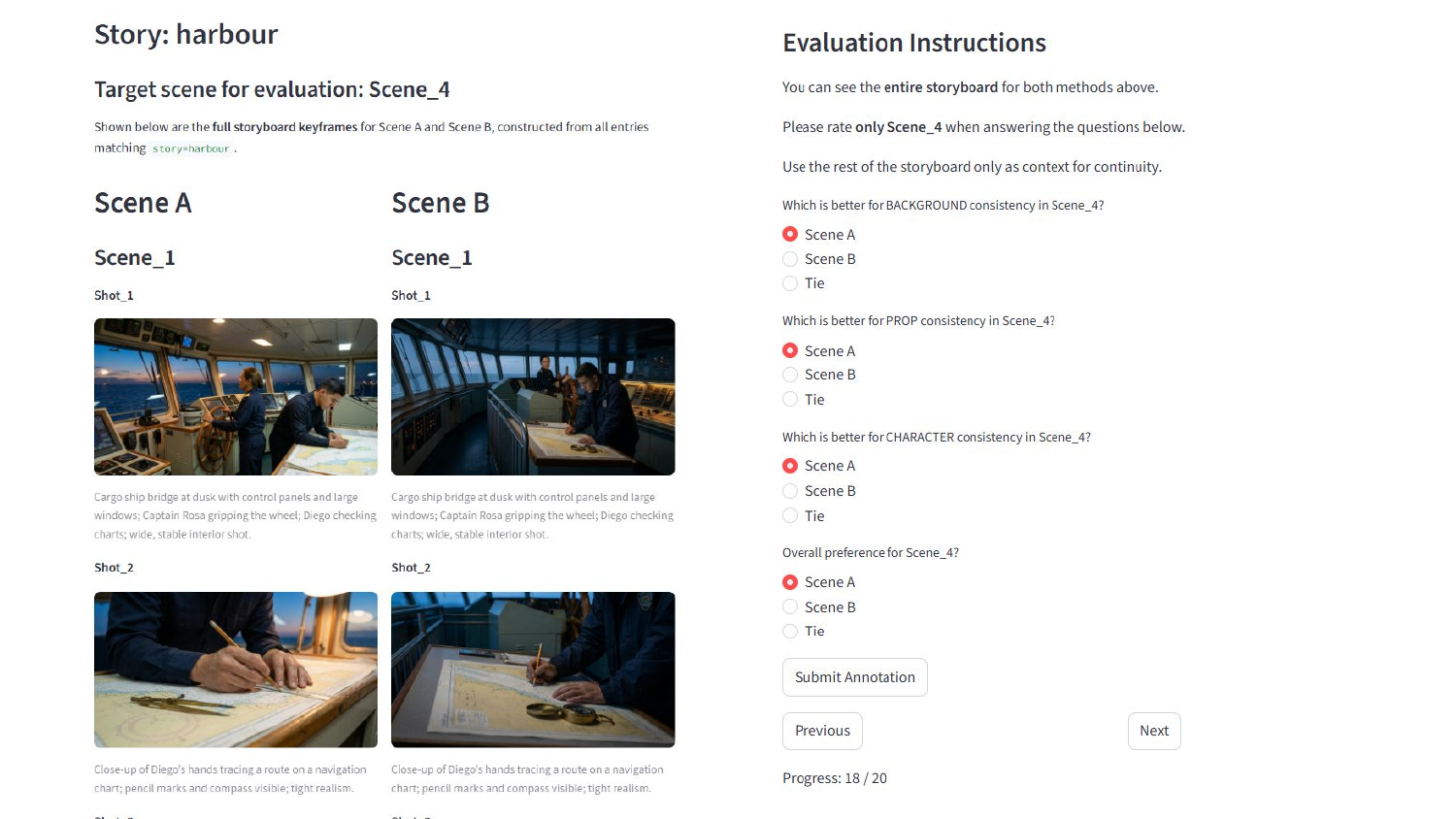}}
    \caption{
Streamlit-Based Interface to Collect Human Preference on the Stories (Shown Side-by-Side) Generated by CANVAS and Other Baselines.}
    \label{fig:humaneval}
\end{figure*}

\begin{table*}[t]
\centering
\tiny
\setlength{\tabcolsep}{4pt}
\renewcommand{\arraystretch}{1.08}
\begin{tabular}{l|cc|c|ccc|c}
\toprule
Method 
& \multicolumn{2}{c|}{Background Consistency}
& Props
& \multicolumn{3}{c|}{Character Consistency}
& Char Avg \\
\cmidrule(lr){2-3}
\cmidrule(lr){5-7}
& Consec & Non-Consec
& PropCons
& FaceCons & ClothCons & BodyCons
& \\
\midrule
AutoStudio \citep{cheng2025autostudiocraftingconsistentsubjects}     & 3.28 & 3.48 & 3.36 & 3.58 & 3.38 & 3.68 & 3.55 \\
Story-Iter \citep{mao2026storyitertrainingfreeiterativeparadigm}      & 3.58 & 3.73 & 3.48 & 3.78 & 3.52 & 3.82 & 3.71 \\
Story2Board  \citep{dinkevich2025story2boardtrainingfreeapproachexpressive}   & 3.12 & 3.32 & 3.02 & 3.50 & 3.28 & 3.52 & 3.43 \\
Gemini-CT \citep{comanici2025gemini25pushingfrontier}  & 4.30 & 3.80 & 4.20 & 3.92 & 3.67 & 3.88 & 3.82 \\
\rowcolor{green!12}
\textbf{CANVAS (Ours)}
& \textbf{4.60} & \textbf{4.60}
& \textbf{4.52}
& \textbf{4.11} & \textbf{4.39} & \textbf{4.02}
& \textbf{4.18} \\
\bottomrule
\end{tabular}
\caption{
ContinuityEval results using GPT-5 as the judge (To avoid the risk of circularity/bias of generation and evaluation using same model). Using GPT-5 as an independent evaluator, CANVAS consistently improves continuity across all dimensions compared to the strongest baseline (Gemini-CT). Background continuity improves from 4.30 → 4.60 (+7.0\%) for consecutive scenes and from 3.80 → 4.60 (+21.1\%) for non-consecutive scene reappearances, indicating substantially stronger long-range environment consistency. Prop persistence also improves from 4.20 → 4.52 (+7.6\%), suggesting that CANVAS better maintains object states across shots.
Character consistency also improves substantially, with the average character score increasing from 3.82 → 4.18 (+9.4\%), indicating more stable identity preservation across shots.
Overall, the results show that CANVAS provides the largest gains in non-consecutive background consistency, highlighting its ability to maintain coherent environments across distant shots—one of the most challenging aspects of long-range visual storytelling.
}
\label{tab:continuityevalwithgtp}
\end{table*}

\begin{table*}[t]
\centering
\small
\setlength{\tabcolsep}{4pt}

\begin{tabular}{l|cc|cc|ccccc|c|c|c|c}
\toprule

Method 
& \multicolumn{2}{c|}{CSD $\uparrow$}
& \multicolumn{2}{c|}{CIDS $\uparrow$}
& \multicolumn{5}{c|}{Prompt Alignment (PA) $\uparrow$}
& CM $\uparrow$
& Inc $\uparrow$
& Aes $\uparrow$
& CP $\uparrow$ \\

\cmidrule(lr){2-3}
\cmidrule(lr){4-5}
\cmidrule(lr){6-10}

& Cross & Self 
& Cross & Self
& Scene & Shot & CI & IA & Avg
& & & & \\

\midrule

Story-Iter      & 0.48 & 0.67 & 0.60 & 0.71 & 3.50 & 3.67 & 3.31 & 3.11 & 3.39 & 57.34 & 14.33 & 4.55 & 0.30 \\
Story2Board     & 0.23 & 0.45 & 0.35 & 0.39 & 3.12 & 2.34 & 3.0 & 3.23 & 2.92 & 44.44 & 11.93 & 4.56 & 0.24 \\
AutoStudio      & 0.34 & 0.63 & 0.44 & 0.46 & 3.34 & 3.55 & 3.56 & 3.11 & 3.39 & 56.22 & 14.34 & 3.45 & 0.235 \\

Gemini-CT    & 0.38 & 0.62 & 0.58 & 0.65 & 3.46 & 3.82 & 3.96 & 3.20 & 3.61 & 59.97 & 12.50 & 5.54 & 0.244 \\

\midrule

\textbf{CANVAS} & \textbf{0.49} & \textbf{0.72} & \textbf{0.61} & \textbf{0.71} & \textbf{3.50} & 3.75 & 3.23 & \textbf{3.9} & 3.60 & \textbf{60.34} & \textbf{14.89} & \textbf{5.60} & \textbf{0.312} \\

\bottomrule
\end{tabular}

\caption{
Comparison on ViStoryBench-Lite across story-image and multimodal generation methods. 
Metrics include Cross-Shot Diversity (CSD), Cross-Image Dependency Score (CIDS), Prompt Alignment (PA), 
Consistency Metric (CM), Inconsistency (Inc), Aesthetic score (Aes), and Character Preservation (CP). 
Gemini-CTo results are taken directly from the benchmark evaluation, and CANVAS uses the same evaluation protocol.
}

\end{table*}

\subsection{Storyboard Quality on Video Generation}
\label{subsubsec:filmEval}
While recent text-to-video systems have made significant progress in generating short clips, producing long-form videos that remain faithful to the intended narrative remains challenging. In practice, many pipelines rely on \textit{storyboard-based generation}, where an image or keyframe guides the initialization of each video segment. However, the quality of these keyframes directly affects the coherence and consistency of the resulting video. If the storyboard fails to preserve character identity, scene layout, or object states across shots, the downstream image-to-video model may amplify these inconsistencies, leading to narrative drift, identity changes, and visually incoherent sequences.

To study the effect of storyboard quality on downstream video generation, we compare different storyboard generation methods by using their keyframes to seed the same image-to-video model. Specifically, we use the Veo-3.1-preview image-to-video model to generate short clips conditioned on the first keyframe produced by each method. All systems use \textit{identical prompts}, ensuring that the only varying factor is the storyboard initialization. The generated clips are concatenated to produce videos of approximately 1.5 minutes, allowing us to evaluate long-range narrative continuity.

Table~\ref{tab:video_eval_results} reports results across eight cinematic evaluation metrics proposed by FilMaster~\citep{huang2025filmasterbridgingcinematicprinciples}, with Gemini-2.5-Flash acting as the automatic judge. The results demonstrate that storyboard generation plays a critical role in determining final video quality. Methods that fail to preserve character identity or scene structure lead to reduced narrative coherence and character consistency in the generated videos. In contrast, \textbf{CANVAS} significantly improves performance across most metrics, achieving the highest overall score and showing particularly strong gains in narrative coherence, character consistency, cinematic techniques, and compelling storytelling. These results highlight the importance of continuity-aware storyboard generation for producing coherent long-form videos.

\begin{table*}[t]
\centering
\tiny
\setlength{\tabcolsep}{6pt}
\begin{tabular}{lcccccccc|c}
\toprule

\textbf{Method}
& \textbf{SF}
& \textbf{NC}
& \textbf{VQ}
& \textbf{CC}
& \textbf{PLC}
& \textbf{CT}
& \textbf{CD}
& \textbf{OQ}
& \textbf{Avg} \\

\midrule

StoryIter \citep{mao2026storyitertrainingfreeiterativeparadigm} + Veo-3.1
& 3.20
& 2.60
& 3.40
& 1.80
& 4.40
& 2.80
& 4.00
& 2.80
& 3.13 \\

AutoStudio \citep{cheng2025autostudiocraftingconsistentsubjects} + Veo-3.1
& 3.00
& 2.80
& 3.20
& 2.00
& 3.50
& 3.00
& 3.50
& 3.20
& 3.03 \\

Gemini-CT \citep{zhang2025storymemmultishotlongvideo} + Veo-3.1
& 3.00
& 2.20
& 3.60
& 2.20
& 3.80
& 3.60
& 3.80
& 4.40
& 3.33 \\

Story2Board \citep{dinkevich2025story2boardtrainingfreeapproachexpressive}
+ Veo-3.1
& 3.60
& 2.20
& 1.80
& 1.00
& 3.60
& 2.20
& 3.40
& 3.20
& 2.63 \\

CANVAS (Ours) + Veo-3.1
& \textbf{3.40}
& \textbf{4.00}
& \textbf{3.80}
& \textbf{4.00}
& \textbf{3.80}
& \textbf{4.00}
& \textbf{4.40}
& \textbf{4.80}
& \textbf{4.03} \\

\bottomrule
\end{tabular}

\caption{
Comparison of the videos generated by each of the baselines and CANVAS paired with a standard I2V model across eight evaluation metrics proposed by FilMaster \citep{huang2025filmasterbridgingcinematicprinciples} using Gemini-2.5-Flash as the judge.
We evaluate videos generated using the Veo-3.1-preview Image-to-Video (I2V) model, where the first frame is initialized with the keyframe produced by each storyboard generation baseline. To ensure fair comparison, we keep the video prompt identical across methods, varying only the keyframe input. Each prompt–keyframe pair is used to generate short clips, which are then concatenated to produce a final video of approximately 1.5 minutes.
Metrics are abbreviated as follows: 
\textbf{SF} (Script Faithfulness), 
\textbf{NC} (Narrative Coherence), 
\textbf{VQ} (Visual Quality), 
\textbf{CC} (Character Consistency), 
\textbf{PLC} (Physical Law Compliance), 
\textbf{CT} (Cinematic Techniques), 
\textbf{CD} (Compelling Degree), and 
\textbf{OQ} (Overall Quality).
\textbf{CANVAS} achieves the strongest overall performance, particularly in narrative coherence, character consistency, cinematic techniques, and compelling storytelling, demonstrating the benefits of continuity-aware generation that maintains consistent characters, environments, and object states across shots.
}
\label{tab:video_eval_results}
\end{table*}

\begin{table*}[t]
\centering
\small
\setlength{\tabcolsep}{4pt}

\begin{tabular}{l|cccccccc}
\toprule
Method 
& SF 
& NC 
& VQ 
& CC 
& PL 
& CT 
& CD 
& OQ \\

\midrule

Story2Board \citep{dinkevich2025story2boardtrainingfreeapproachexpressive}
+ Veo-3.1 & 2.25 & 1.75 & 1.75 & 1.00 & 3.00 & 2.75 & 2.75 & 2.25 \\
Story-Iter \citep{mao2026storyitertrainingfreeiterativeparadigm} + Veo-3.1  & 3.75 & 2.25 & 3.75 & 2.00 & 3.75 & 4.50 & 4.25 & 4.25 \\
Gemini-CT + + Veo-3.1      & 3.75 & 2.25 & 4.25 & 2.50 & 3.50 & 4.25 & 4.00 & 4.00 \\

\rowcolor{gray!10}
CANVAS (Ours) + Veo-3.1 & \textbf{4.00} & \textbf{3.00} & \textbf{4.75} & \textbf{4.00} & 3.25 & \textbf{5.00} & \textbf{4.75} & \textbf{4.75} \\

\bottomrule
\end{tabular}

\caption{Comparison of the videos generated by each of the baselines and CANVAS paired with a standard I2V model across eight evaluation metrics proposed by FilMaster \citep{huang2025filmasterbridgingcinematicprinciples} using Gemini-2.5-Pro as the judge.
We evaluate videos generated using the Veo-3.1-preview Image-to-Video (I2V) model, where the first frame is initialized with the keyframe produced by each storyboard generation baseline. To ensure fair comparison, we keep the video prompt identical across methods, varying only the keyframe input. Each prompt–keyframe pair is used to generate short clips, which are then concatenated to produce a final video of approximately 1.5 minutes.
Metrics are abbreviated as follows: 
\textbf{SF} (Script Faithfulness), 
\textbf{NC} (Narrative Coherence), 
\textbf{VQ} (Visual Quality), 
\textbf{CC} (Character Consistency), 
\textbf{PLC} (Physical Law Compliance), 
\textbf{CT} (Cinematic Techniques), 
\textbf{CD} (Compelling Degree), and 
\textbf{OQ} (Overall Quality).
\textbf{CANVAS} achieves the strongest overall performance, particularly in narrative coherence, character consistency, cinematic techniques, and compelling storytelling, demonstrating the benefits of continuity-aware generation that maintains consistent characters, environments, and object states across shots.}
\end{table*}

\begin{table*}[t]
\centering
\tiny
\setlength{\tabcolsep}{4pt}

\begin{tabular}{l l | cc cc | cccc c | ccc c}
\toprule

Method & Model 
& \multicolumn{2}{c}{CSD$\uparrow$} 
& \multicolumn{2}{c}{CIDS$\uparrow$}
& \multicolumn{5}{c}{PA$\uparrow$}
& CM$\uparrow$ & Inc$\uparrow$ & Aes$\uparrow$ & CP \\

\cmidrule(lr){3-4}
\cmidrule(lr){5-6}
\cmidrule(lr){7-11}

& 
& Cross & Self 
& Cross & Self
& Scene & Shot & CI & IA & Avg.
& & & & \\

\midrule

Copy-Paste Baseline & - & 0.735 & 0.770 & 0.911 & 0.993 & 0.62 & 2.08 & 0.87 & 1.55 & 1.28 & 92.76 & 5.46 & 4.39 & 0.550 \\

\midrule
\multicolumn{14}{c}{\textbf{Story Image Method}} \\
\midrule

StoryGen \citep{Liu_2024_CVPR} & SD1.5 & 0.405 & 0.562 & 0.405 & 0.591 & 1.05 & 2.32 & 1.43 & 1.75 & 1.64 & 52.98 & 7.15 & 4.09 & 0.277 \\
StoryGen \citep{Liu_2024_CVPR} & SD1.5 & 0.396 & 0.551 & 0.396 & 0.602 & 1.18 & 2.25 & 1.49 & 1.50 & 1.60 & 52.53 & 7.67 & 4.09 & 0.224 \\
StoryGen \citep{Liu_2024_CVPR} & SD1.5 & 0.316 & 0.617 & 0.316 & 0.610 & 0.89 & 2.41 & 1.39 & 2.05 & 1.68 & 40.13 & 6.25 & 3.86 & 0.240 \\
TheaterGen \citep{cheng2024theatergen} & SD1.5 & 0.221 & 0.411 & 0.354 & 0.537 & 2.85 & 1.94 & 1.02 & 1.15 & 1.74 & 54.93 & 13.60 & 4.94 & 0.204 \\
StoryDiffusion \citep{zhou2024storydiffusionconsistentselfattentionlongrange} & SDXL & 0.293 & 0.680 & 0.409 & 0.641 & 3.21 & 3.00 & 1.92 & 1.95 & 2.52 & 67.07 & 12.99 & 5.83 & 0.186 \\
StoryDiffusion \citep{zhou2024storydiffusionconsistentselfattentionlongrange} & SDXL & 0.409 & 0.611 & 0.460 & 0.575 & 1.91 & 3.41 & 2.64 & 2.75 & 2.68 & 62.48 & 8.18 & 5.21 & 0.251 \\
SEED-Story \citep{yang2024seedstory} & SDXL & 0.258 & 0.763 & 0.559 & 0.656 & 2.17 & 1.93 & 1.49 & 2.89 & 1.87 & 64.33 & 4.90 & 3.81 & 0.306 \\
Story-Adapter \citep{mao2026storyitertrainingfreeiterativeparadigm} & SD1.5 & 0.518 & 0.609 & 0.490 & 0.605 & 1.91 & 3.44 & 2.56 & 2.70 & 2.65 & 70.34 & 11.49 & 4.89 & 0.250 \\
Story-Adapter \citep{mao2026storyitertrainingfreeiterativeparadigm} & SD1.5 & 0.371 & 0.758 & 0.425 & 0.619 & 1.80 & 3.20 & 2.38 & 2.30 & 2.42 & 61.39 & 12.03 & 4.80 & 0.217 \\
Story-Adapter \citep{mao2026storyitertrainingfreeiterativeparadigm} & SD1.5 & 0.343 & 0.515 & 0.430 & 0.547 & 1.97 & 3.42 & 2.74 & 2.65 & 2.69 & 65.32 & 12.72 & 5.12 & 0.203 \\
Story-Adapter \citep{mao2026storyitertrainingfreeiterativeparadigm} & SD1.5 & 0.353 & 0.752 & 0.416 & 0.634 & 1.85 & 3.23 & 2.48 & 2.30 & 2.46 & 61.57 & 10.59 & 4.85 & 0.220 \\
UNO \citep{uno} & FLUX1 & 0.425 & 0.648 & 0.512 & 0.630 & 3.54 & 2.98 & 3.54 & 2.55 & 3.06 & 70.88 & 10.50 & 5.13 & 0.287 \\
OmniGen2 \citep{wu2025omnigen2explorationadvancedmultimodal} & DiT & 0.491 & 0.648 & 0.576 & 0.668 & 3.49 & 3.29 & 3.49 & 2.55 & 3.07 & 73.44 & 8.21 & 5.21 & 0.298 \\
CharaConsist \citep{wang2025characonsistfinegrainedconsistentcharacter} & FLUX1 & 0.333 & 0.646 & 0.347 & 0.539 & 3.69 & 3.61 & 2.86 & 2.21 & 3.09 & 62.10 & 10.84 & 5.78 & 0.216 \\
QwenImageEdit-2509 \citep{wu2025qwenimagetechnicalreport} & DiT & 0.404 & 0.614 & 0.482 & 0.541 & 3.89 & 3.53 & 3.89 & 2.60 & 3.25 & 61.27 & 10.56 & 5.46 & 0.249 \\
\midrule
\multicolumn{14}{c}{\textbf{Multi-modal Large Model (Language, Image and Video)}} \\
\midrule

GPT-4o* & - & 0.481 & 0.680 & 0.420 & 0.522 & 3.649 & 3.781 & 3.907 & 2.684 & 3.505 & 69.33 & 9.02 & 5.49 & 0.209 \\
Gemini-2.0* & - & 0.361 & 0.573 & 0.573 & 0.677 & 3.261 & 3.565 & 3.198 & 3.294 & 3.329 & 74.82 & 10.12 & 4.91 & 0.266 \\
Gemini-2.5* & - & 0.447 & 0.657 & 0.553 & 0.642 & 3.319 & 3.709 & 3.991 & 3.250 & 3.567 & 64.86 & 10.54 & 5.61 & 0.255 \\
Gemini-3.0 Pro*  & - & 0.385 & 0.622 & 0.581 & 0.653 & 3.458 & 3.828 & 3.969 & 3.200 & 3.614 & 59.97 & 12.50 & 5.54 & 0.244 \\

\midrule
\multicolumn{13}{c}{\textbf{Our Method}} \\

\midrule

CANVAS & Gemini-3-pro &
 0.491 &  0.72 & 0.61 & 0.71 & 3.5 & 3.75 & 3.23 & 3.9 & 3.6 & 60.34 & 14.89 & 5.60 & 0.312\\
\bottomrule
\end{tabular}

\caption{Benchmark comparison across story image and multimodal large models on ViStoryBench-Lite \citep{zhuang2025vistorybenchcomprehensivebenchmarksuite}. We copy the table from the paper to report numbers of the existing baselines (Only Story-Image and Open-source Models excluding the StoryVideo methods and Commercial Platforms) and compare CANVAS using the same evaluation suite released by the paper.}
\end{table*}

\begin{figure*}[t]
\centering

\begin{tabular}{p{0.47\linewidth} p{0.47\linewidth}}

\textbf{Shot 1.}
Inside a warmly lit wooden workshop at the North Pole, Santa Claus—an elderly man with a white beard and red suit—sits at a sturdy table writing names in a large leather-bound book. Toys and tools fill the background.
&
\textbf{Shot 2.}
Close-up of Santa gently polishing a small wooden toy train, smiling as he tests its wheels in the cozy workshop environment. \\[4pt]

\includegraphics[width=\linewidth]{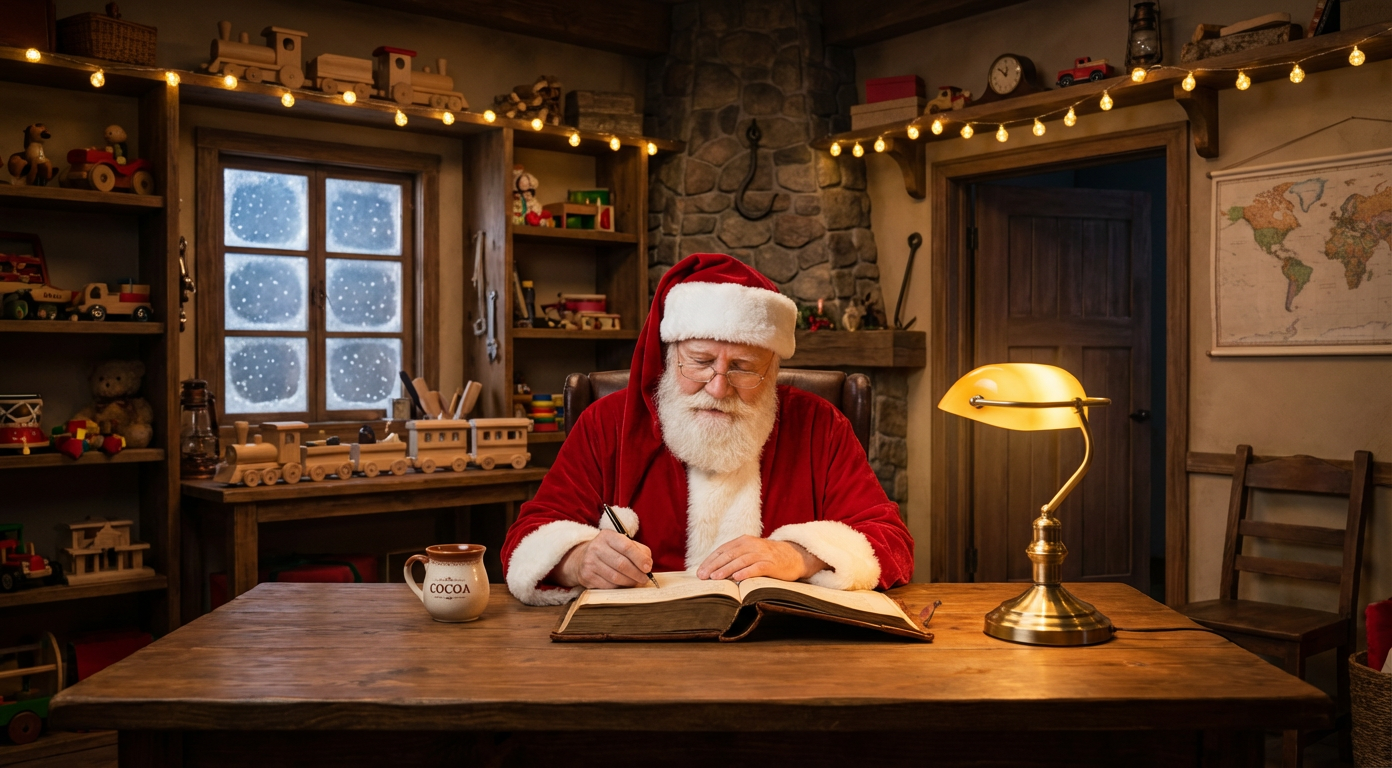} &
\includegraphics[width=\linewidth]{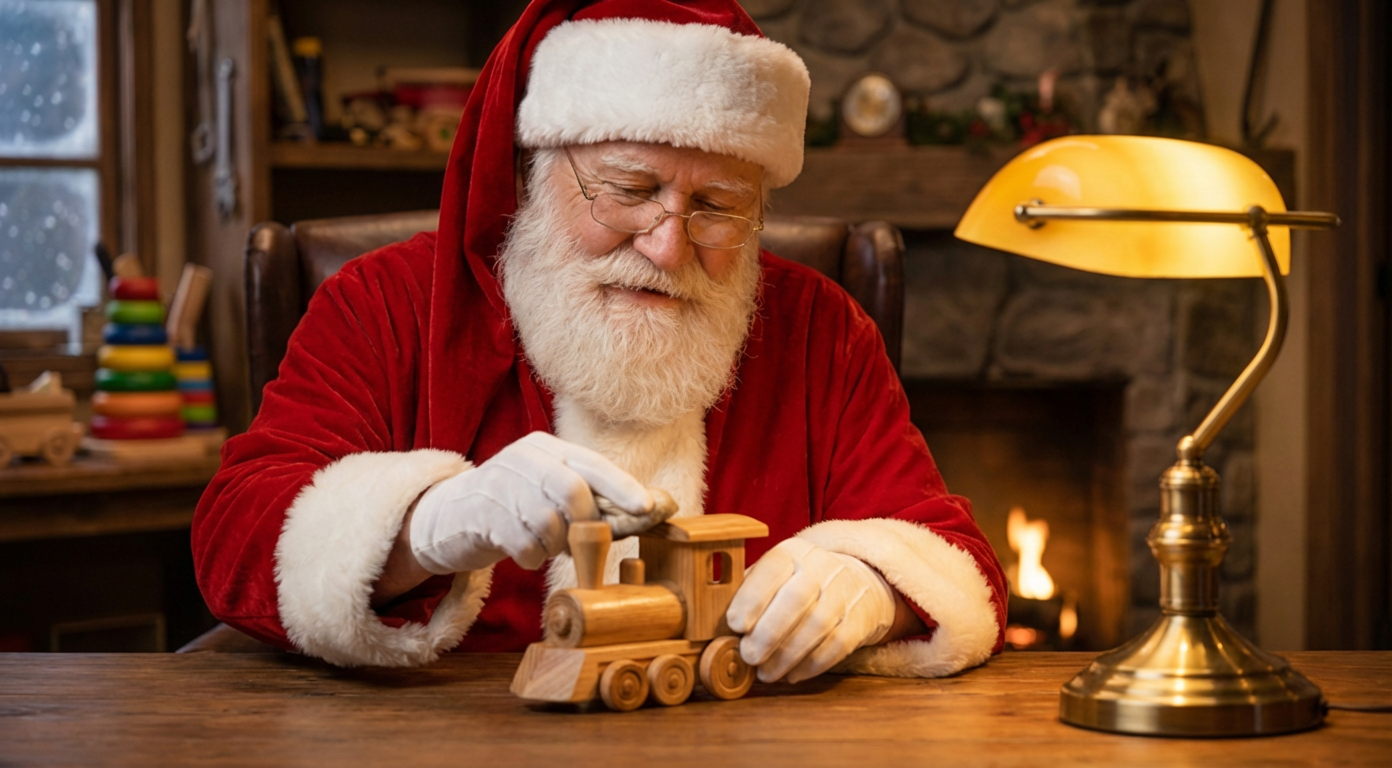} \\

\small (a) CANVAS (w/ BackgroundPlanning) &
\small (b) CANVAS (w/ Background Planning) \\

\includegraphics[width=\linewidth]{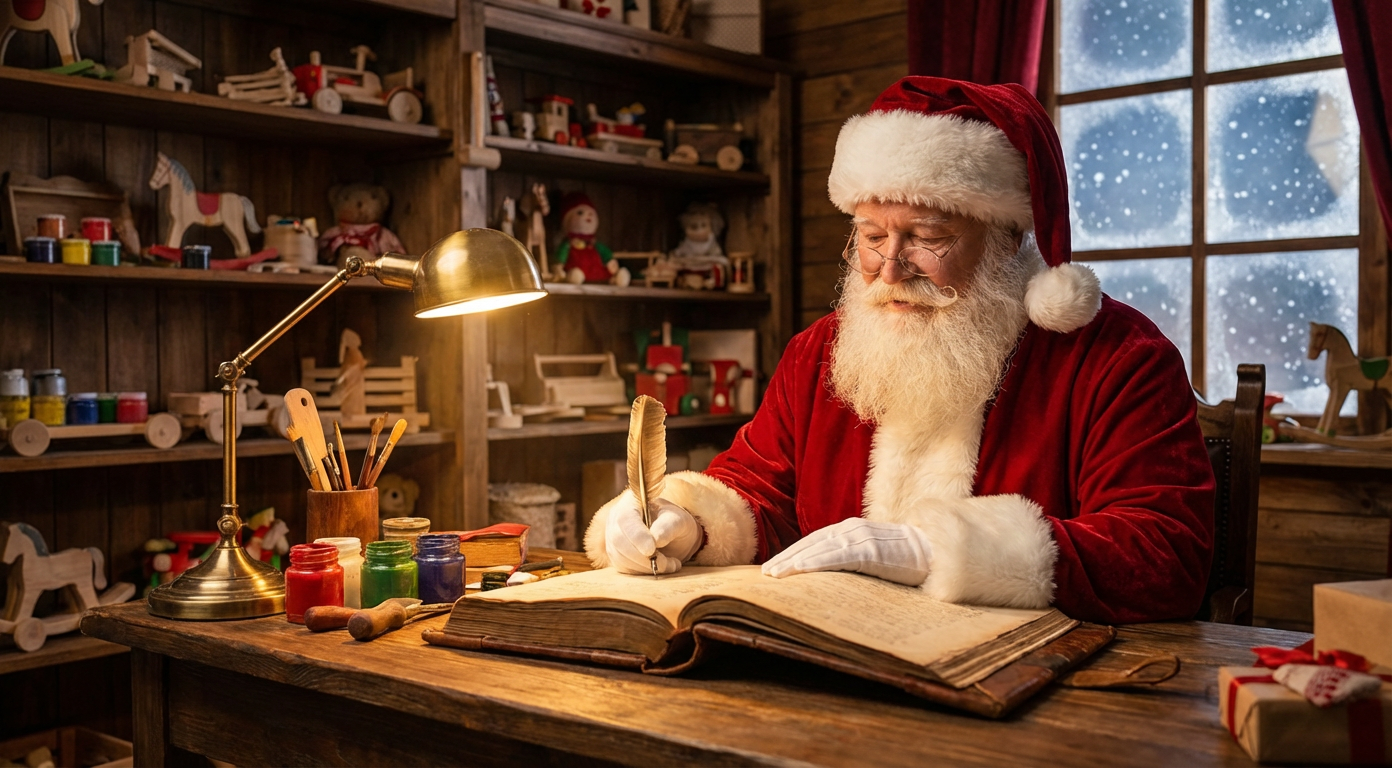} &
\includegraphics[width=\linewidth]{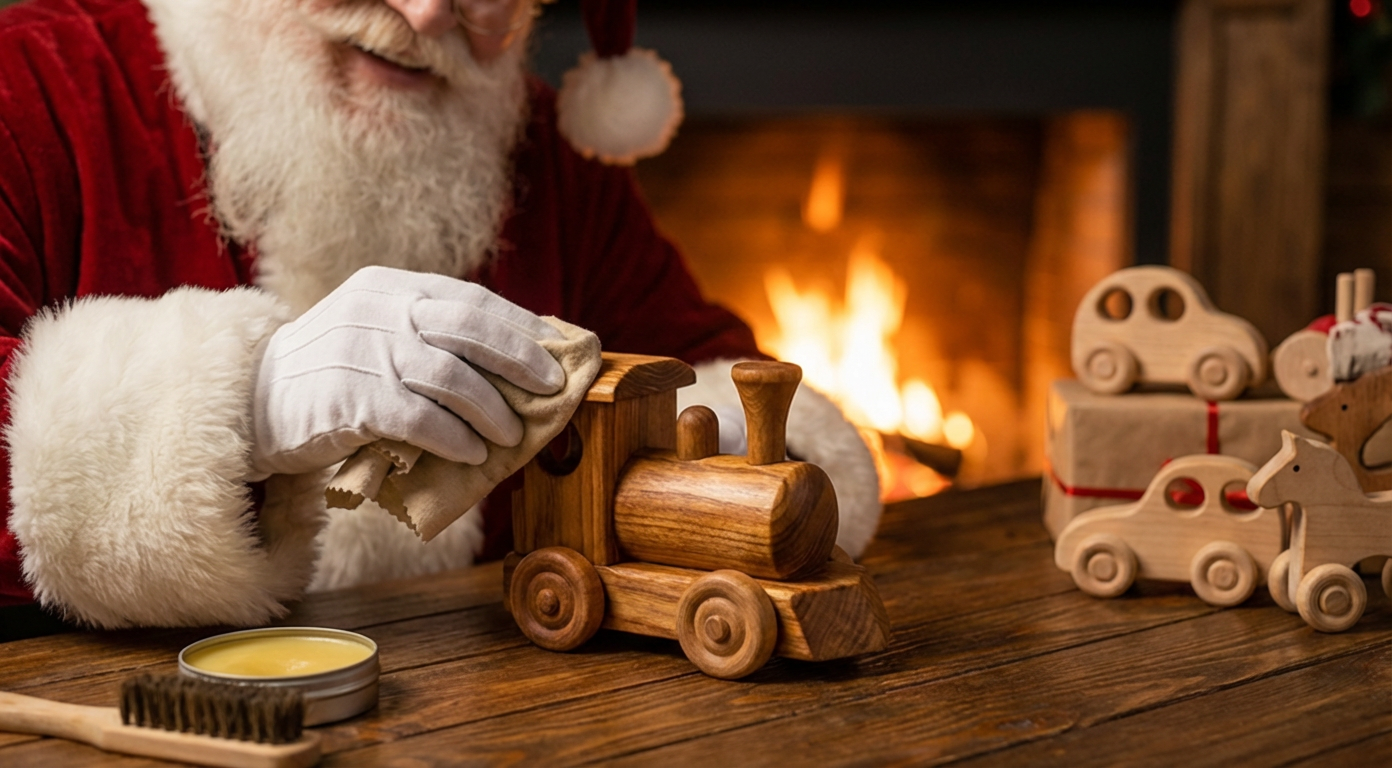} \\

\small (c) CANVAS (w/o Background Planning) &
\small (d) CANVAS (w/o Background Planning) \\

\end{tabular}

\caption{
Qualitative comparison of cross-shot continuity with and without background Planning.
The first row shows outputs from \textbf{CANVAS with structured planning}, which performs location canonicalization and prop forecasting before shot generation.
The second row shows results \textbf{without background planning}.
While both depict the same narrative interaction, in the first shot, the fireplace remains occluded and also the toy was present at the background. Hence, the next shot just looks like a zoomed-in version of Santa, maintaining higher narrative continuity.\\ 
}

\label{fig:shot_comparison}

\end{figure*}

\begin{figure*}[!t]
    \centering
    \includegraphics[width=\linewidth]{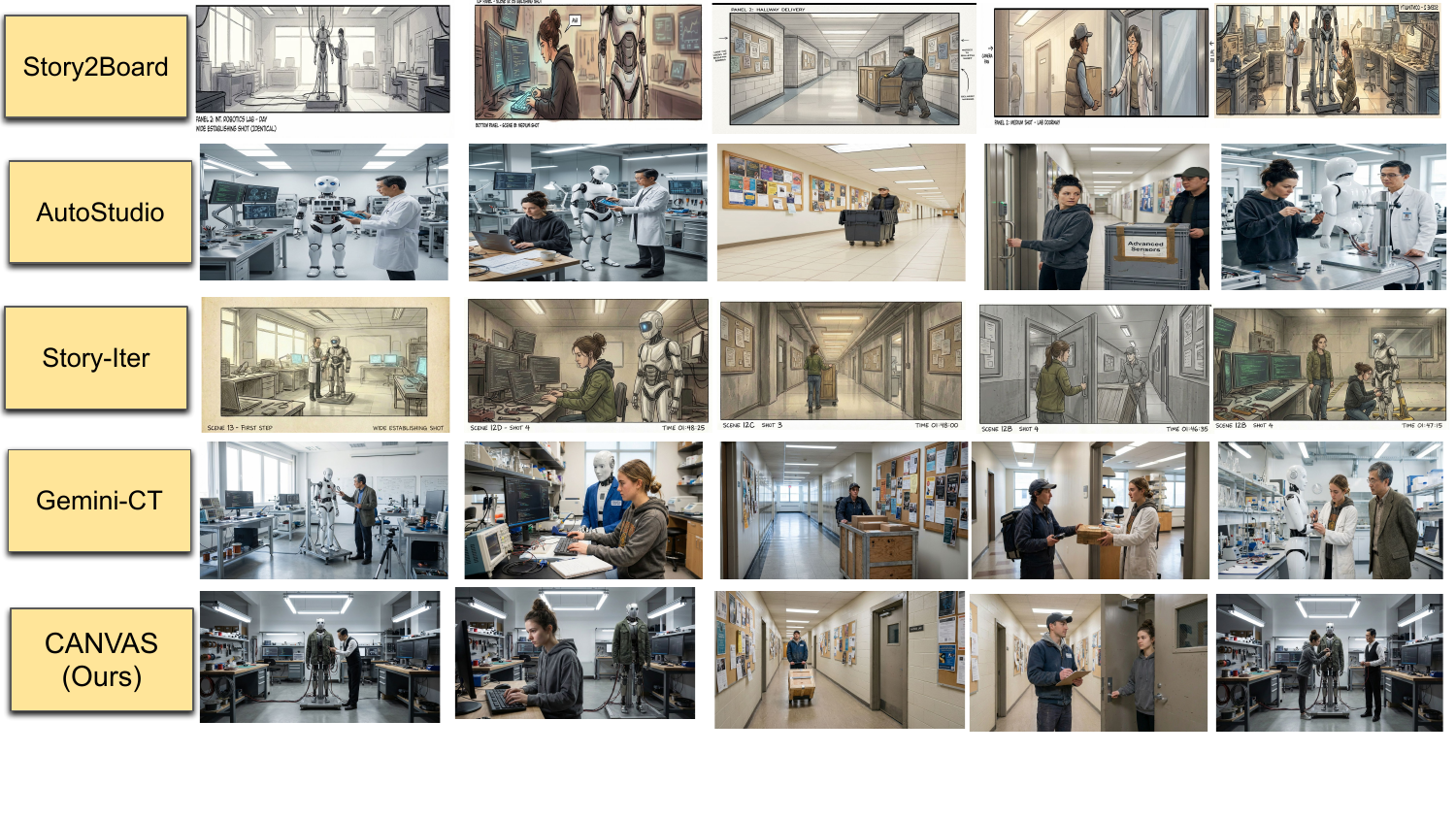}
    \caption{The figure compares storyboards generated by recent training-free methods—Story2Board, AutoStudio, Story-Iter, Gemini-CT—and CANVAS for a multi-shot narrative set across a robotics lab and hallway. Each row shows a method and each column shows a shot, illustrating how well character identity, environments, and object states are maintained across the sequence. Existing methods produce visually plausible frames but suffer from identity drift, layout inconsistencies, and unstable object states, especially when scenes reappear after transitions. In contrast, CANVAS preserves consistent character appearance, stable environments, and coherent object interactions throughout the narrative by using global continuity planning and world-state memory tracking. Overall, the figure shows that explicitly modeling characters, locations, and object states enables CANVAS to maintain strong long-range visual coherence across multi-shot storyboards.
}
    \label{fig:labexample}
\end{figure*}

\subsection{Prompts and Implementation Details for CANVAS}
\label{app:canvas}

\begin{table*}[t]
\centering
\setlength{\fboxsep}{6pt}

\begin{tabular}{|p{0.95\linewidth}|}
\hline
\rowcolor{blue!15}
\textbf{Prompt: Location Clustering for Background Continuity} \\ \hline

\rowcolor{blue!3}
You are a narrative continuity planner for visual storytelling. 
Your goal is to identify and cluster recurring locations across a sequence of storyboard shots. \\

\textbf{Instructions:}
\begin{enumerate}
\item Identify the physical environment in which each shot occurs.
\item Assign a unique \texttt{location\_id} to each distinct location.
\item Cluster shots that occur in the same environment under the same \texttt{location\_id}.
\item If a shot returns to a previously seen environment, mark it as \texttt{location\_reappearance}.
\item If a shot directly continues the previous scene without changing environment, mark it as \texttt{previous\_frame\_continuation}.
\item If the location appears for the first time, mark it as \texttt{fresh\_location}.
\end{enumerate}

\textbf{Return STRICT JSON in the following format:}

\begin{verbatim}
{
 "locations": {
   "museum_gallery": [
     "Scene_1_Shot_1",
     "Scene_1_Shot_2",
     "Scene_3_Shot_1"
   ],
   "security_room": [
     "Scene_2_Shot_1"
   ]
 },
 "shots": {
   "Scene_1_Shot_1": {
     "location_id": "museum_gallery",
     "continuity_mode": "fresh_location"
   }
 }
}
\end{verbatim}

\textbf{Constraints}
\begin{itemize}
\item Maintain consistent location identities across shots.
\item Do not create duplicate location IDs for the same environment.
\item When a location reappears later in the story, reuse the original \texttt{location\_id}.
\end{itemize}

\\ \hline
\end{tabular}

\caption{Prompt used for clustering shots into recurring locations to maintain background continuity across the storyboard.}
\label{tab:location_clustering_prompt}

\end{table*}

\begin{table*}[t]
\centering
\small

\begin{tabular}{|p{0.95\linewidth}|}
\hline
\rowcolor{blue!15}
\textbf{Prompt: Future Prop and Geometry Aware Background Planning} \\ \hline

\rowcolor{purple!3}
You are a story continuity planner responsible for maintaining consistent environments across storyboard shots. Your task is to reason about how the background scene should evolve based on the current and future states of important props and architectural geometry of the background location. \\

\textbf{Instructions}

Given:
\begin{itemize}
\item The description of the current shot.
\item The planned metadata for the shot (characters, location, prop states).
\item The previously known world state of props and environments.
\end{itemize}

Plan the background scene so that it remains logically consistent with both the current story state and upcoming events.

\textbf{Reason about the following:}

\begin{enumerate}
\item Which props should remain visible in the background environment.
\item Which props should disappear because they were taken, destroyed, or moved.
\item Which props are likely to appear in the background due to upcoming story events.
\item Which props are currently carried by characters and therefore should not remain in the environment.
\item Which environmental objects must persist across shots to maintain scene identity (e.g., furniture, structures, display cases).
\item Whether the camera framing hides some props even though they still exist in the environment.
\end{enumerate}

\textbf{Return STRICT JSON in the following format:}

\begin{verbatim}
{
 "background_props": [
   "display_case",
   "museum_sign",
   "security_camera"
 ],
 "must_appear": [
   "golden_artifact"
 ],
 "must_not_appear": [
   "artifact_case_glass"
 ],
 "carried_props": {
   "thief": ["golden_artifact"]
 },
 "reasoning": "The artifact is carried away by the thief,
 so it should not remain in the gallery background.
 The broken display case should remain visible."
}
\end{verbatim}

\textbf{Constraints}

\begin{itemize}
\item Background elements should remain consistent with previously established environments.
\item If a prop was present earlier in the same location and nothing removed it, it should persist.
\item Props carried by characters must not remain in the background.
\item The planned background should support future story events.
\end{itemize}

\\ \hline
\end{tabular}

\caption{Prompt used for future prop-aware background planning to ensure that scene environments evolve consistently with object state transitions.}

\label{tab:future_prop_background_prompt}

\end{table*}

\begin{table*}[t]
\centering

\begin{tabular}{|p{0.95\linewidth}|}
\hline
\rowcolor{blue!15}
\textbf{Prompt: Character Appearance State Planning (Single Character)} \\ \hline

You are a narrative continuity planner responsible for tracking the visual appearance of a single character across a sequence of storyboard shots. Your goal is to determine the appearance state of the character in each shot and ensure that the appearance evolves consistently with the story. \\

\textbf{Input}

\begin{itemize}
\item The name of the target character.
\item A sequence of storyboard shot descriptions.
\end{itemize}

\textbf{Instructions}

\begin{enumerate}
\item For each shot, determine whether the target character appears in the scene.
\item If the character appears, determine the character's visual appearance state (e.g., clothing, uniform, disguise).
\item If the appearance does not change, reuse the same appearance state identifier as the previous shot.
\item If the story explicitly changes the character’s clothing or visual style, assign a new appearance state.
\item If the character does not appear in a shot, mark the state as \texttt{not\_present}.
\end{enumerate}

\textbf{Return STRICT JSON in the following format:}

\begin{verbatim}
{
 "character": "Ethan",
 "appearance_timeline": {
   "Scene_1_Shot_1": "blue_curator_jacket",
   "Scene_1_Shot_2": "blue_curator_jacket",
   "Scene_1_Shot_3": "blue_curator_jacket",
   "Scene_2_Shot_1": "black_security_jacket",
   "Scene_2_Shot_2": "not_present",
   "Scene_3_Shot_1": "black_security_jacket"
 }
}
\end{verbatim}

\textbf{Constraints}

\begin{itemize}
\item Maintain consistent appearance across shots unless the story explicitly indicates a change.
\item Reuse appearance identifiers whenever the clothing or visual style remains the same.
\item Ensure that later reappearances match the most recent appearance state.
\end{itemize}

\\ \hline
\end{tabular}

\caption{Prompt used during Character Appearance Planning.}
\label{tab:character_appearance_single}

\end{table*}

\begin{table*}[t]
\centering
\small

\begin{tabular}{|p{0.95\linewidth}|}
\hline
\rowcolor{blue!15}
\textbf{Prompt: Prop State Update Planning (Single Object)} \\ \hline

You are a narrative continuity planner responsible for tracking the state of a single prop (object) across a sequence of storyboard shots. Your goal is to determine how the object's state evolves throughout the story and ensure logical consistency with the described events. \\

\textbf{Input}

\begin{itemize}
\item The name of the target prop (object).
\item A sequence of storyboard shot descriptions.
\end{itemize}

\textbf{Instructions}

\begin{enumerate}
\item For each shot, determine whether the target prop appears in the scene.
\item If the prop appears, determine its state (e.g., intact, broken, inside container, carried by character, missing).
\item If the prop is carried by a character, record which character carries it.
\item If the prop is removed from the scene (e.g., stolen, destroyed, moved), update its state accordingly.
\item If the prop continues unchanged across shots, reuse the same state identifier.
\item If the prop does not appear in a shot but still exists in the story world, mark it as \texttt{not\_visible}.
\end{enumerate}

\textbf{Return STRICT JSON in the following format:}

\begin{verbatim}
{
 "prop": "golden_artifact",
 "state_timeline": {
   "Scene_1_Shot_1": {
      "state": "inside_display_case",
      "carrier": null
   },
   "Scene_1_Shot_2": {
      "state": "inside_display_case",
      "carrier": null
   },
   "Scene_1_Shot_3": {
      "state": "carried",
      "carrier": "masked_thief"
   },
   "Scene_2_Shot_1": {
      "state": "not_visible",
      "carrier": "masked_thief"
   },
   "Scene_3_Shot_1": {
      "state": "missing_from_gallery",
      "carrier": "masked_thief"
   }
 }
}
\end{verbatim}

\textbf{Constraints}

\begin{itemize}
\item Maintain consistent prop states across shots unless the story explicitly changes them.
\item If a prop is carried by a character, it should not remain in the previous location.
\item If a prop disappears from the environment, update its state to reflect the narrative event.
\item When the prop reappears later, ensure its state matches the last known world state.
\end{itemize}

\\ \hline
\end{tabular}

\caption{Prompt used during global planning to determine the state timeline for a single prop across shots.}
\label{tab:prop_state_planning_prompt}

\end{table*}

\begin{table*}[t]
\centering
\tiny

\begin{tabular}{|p{0.95\linewidth}|}
\hline
\rowcolor{blue!15}
\textbf{Prompt: Previous-Frame Continuation Decision} \\ \hline

You are a visual continuity planner for storyboard generation. Your task is to decide whether the current shot is a direct continuation of the previous shot, such that the previous frame should be reused as a visual anchor for generation. \\

\textbf{Input}

\begin{itemize}
\item The description of the previous shot.
\item The description of the current shot.
\item The location assignment of both shots.
\item The characters present in both shots.
\item The prop states in both shots.
\end{itemize}

\textbf{Instructions}

Determine whether the current shot should be treated as \texttt{previous\_frame\_continuation}, \texttt{location\_reappearance}, or \texttt{fresh\_location}.

Mark the current shot as \texttt{previous\_frame\_continuation} only if the current shot depends on preserving the spatial structure of the immediately previous frame.

Reason about the following:

\begin{enumerate}
\item Whether the previous shot and current shot occur in the same physical environment.
\item Whether the current shot is an immediate temporal continuation rather than a later revisit.
\item Whether the spatial arrangement of the scene should remain consistent across the cut.
\item Whether important background geometry must be preserved, such as walls, doors, furniture, display cases, tables, platforms, or room layout.
\item Whether character positions, object placements, or scene composition depend on the previous frame.
\item Whether the current shot is a close-up, zoom-in, alternate camera angle, or tighter crop of the same ongoing scene.
\item Whether there are state changes in the scene that still require maintaining the same base spatial environment.
\end{enumerate}

\textbf{Use the following rules:}

\begin{itemize}
\item Choose \texttt{previous\_frame\_continuation} if the scene is still unfolding in the same environment and preserving the exact spatial setup from the previous frame is important.
\item Choose \texttt{location\_reappearance} if the shot returns to a known location after an intervening scene or temporal gap, where the same location identity should be preserved but not necessarily the exact previous composition.
\item Choose \texttt{fresh\_location} if the shot occurs in a new environment not previously shown.
\item Even if some props or characters change state, still choose \texttt{previous\_frame\_continuation} when the underlying scene geometry should remain continuous.
\end{itemize}

\textbf{Return STRICT JSON in the following format:}

\begin{verbatim}
{
  "continuity_mode": "previous_frame_continuation",
  "same_location": true,
  "requires_spatial_continuity": true,
  "spatial_dependencies": [
    "display_case remains in same position",
    "same museum gallery layout",
    "artifact remains centered under spotlight"
  ],
  "reasoning": "..."
}
\end{verbatim}

\textbf{Constraints}

\begin{itemize}
\item Base the decision on scene continuity, not only on shared characters or shared location name.
\item Prioritize whether preserving spatial arrangement from the immediately previous frame will improve visual continuity.
\item If the current shot is compositionally dependent on the previous frame, prefer \texttt{previous\_frame\_continuation}.
\end{itemize}

\\ \hline
\end{tabular}

\caption{Prompt used to decide whether the current shot should reuse the previous frame as a continuation anchor.}
\label{tab:prev_frame_continuation_prompt}

\end{table*}

\begin{algorithm*}[t]
\caption{Candidate Anchor Retrieval and Prompt Construction}
\label{alg:anchor_retrieval}
\tiny
\textbf{Input:} 
Current shot $s_t$, previous frame $I_{t-1}$, global plan $P$, 
character memory $M_c$, location memory $M_l$, prop memory $M_o$

\textbf{Output:}
Candidate generation prompt $\Pi_t$ and anchor set $\mathcal{A}_t$

\begin{enumerate}

\item Initialize anchor set: $A_t \leftarrow \emptyset$

\item Retrieve shot plan from global planner:
\[
p_t = P(s_t)
\]

\item \textbf{Continuation Reasoning}

Determine whether the current shot is a direct continuation of the previous shot.

\begin{enumerate}
\item If the planner determines that spatial dependencies from the previous frame must be preserved:
\begin{enumerate}
\item Add most recent frame anchor:
\[
A_t \leftarrow A_t \cup \{I_{t-1}\}
\]
\item Proceed to character anchor retrieval
\end{enumerate}
\end{enumerate}

\item \textbf{Character Anchor Retrieval}

For each character $c$ appearing in shot $s_t$:

\begin{enumerate}
\item Retrieve appearance state from global plan:
\[
k = C(c,t)
\]

\item If appearance state changed from the previous shot:
\begin{enumerate}
\item Retrieve canonical anchor from canonical memory
\[
I_c = M_c^{canonical}(c,k)
\]
\end{enumerate}

\item Else if appearance state remains unchanged:
\begin{enumerate}
\item Retrieve most recent anchor for the character
\[
I_c = M_c^{recent}(c)
\]
\item If no recent anchor exists:
\[
I_c = M_c^{canonical}(c,k)
\]
\end{enumerate}

\item Add character anchor to anchor set:
\[
A_t \leftarrow A_t \cup \{I_c\}
\]

\end{enumerate}

\item \textbf{Background Anchor Retrieval}

\begin{enumerate}
\item Retrieve location identifier:
\[
\ell = L(s_t)
\]

\item Use prop-state reasoning $O(o,t)$ and background reasoning to determine which background anchor should be retrieved.

\item If location $\ell$ exists in location memory:
\begin{enumerate}
\item Retrieve stored background anchor:
\[
I_\ell = M_l(\ell)
\]
\item Add to anchor set:
\[
A_t \leftarrow A_t \cup \{I_\ell\}
\]
\end{enumerate}

\item Else (first occurrence of the location):
\begin{enumerate}
\item No background anchor is retrieved
\item Scene will be generated using the shot description and prop reasoning
\end{enumerate}

\end{enumerate}

\item \textbf{Prop Anchor Retrieval}

For each prop $o$ appearing in shot $s_t$:

\begin{enumerate}

\item Retrieve prop state from global plan:
\[
s = O(o,t)
\]

\item If prop has appeared previously in the story:
\begin{enumerate}
\item Retrieve stored prop anchor:
\[
I_o = M_o(o)
\]
\item Add to anchor set:
\[
A_t \leftarrow A_t \cup \{I_o\}
\]
\end{enumerate}

\end{enumerate}

\item \textbf{Candidate Prompt Construction}

Construct the candidate generation prompt $\Pi_t$ using:

\begin{enumerate}

\item Shot description $s_t$
\item Retrieved anchors $\mathcal{A}_t$
\item Character appearance states $C(c,t)$
\item Prop states $O(o,t)$
\item Background reasoning from location plan $L(s_t)$

\end{enumerate}

\item Return $(A_t, \Pi_t)$

\end{enumerate}
\label{algo:retrieval}
\end{algorithm*}

\begin{table*}[t]
\centering
\small

\begin{tabular}{|p{0.95\linewidth}|}
\hline
\rowcolor{blue!15}
\textbf{Prompt: Candidate Frame Generation} \\ \hline

\rowcolor{blue!3}
You are a cinematic storyboard generator. Your task is to generate a candidate frame for the current shot while maintaining visual continuity with previously generated frames and respecting the global story plan. \\

\textbf{Input}

\begin{itemize}
\item Shot description $s_t$
\item Retrieved visual anchors $\mathcal{A}_t$ (may include previous frame, character anchors, background anchors, and prop anchors)
\item Character appearance states from the global plan
\item Prop states and prop reasoning for the current shot
\item Location identifier and background reasoning
\end{itemize}

\textbf{Instructions}

Generate a visually coherent frame for the current shot by following these rules:

\begin{enumerate}

\item \textbf{Respect the shot description}

Ensure the generated frame faithfully depicts the actions, characters, and environment described in the current shot.

\item \textbf{Preserve character identity}

Characters must maintain the appearance state specified by the global plan.  
If character anchors are provided, match the visual identity (face, clothing, body appearance) shown in those anchors.

\item \textbf{Maintain spatial continuity}

If the previous frame is provided as an anchor, treat the current shot as a continuation of the same scene.  
Preserve the spatial layout of the environment, including background structure, object placement, and character positions.

\item \textbf{Use background anchors}

If background anchors are provided, maintain the same environment geometry (e.g., walls, furniture, display cases, room layout).

\item \textbf{Respect prop states}

Ensure that props appear according to the planned prop states:

\begin{itemize}
\item Props that must appear should be clearly visible.
\item Props carried by characters should move with the corresponding character.
\item Props that were removed, destroyed, or taken away should not remain in the environment.
\end{itemize}

\item \textbf{Maintain cinematic coherence}

The frame should resemble a realistic cinematic shot with consistent lighting, perspective, and scene composition.

\end{enumerate}

\textbf{Output}

Generate a high-quality candidate frame that satisfies the shot description and respects all provided anchors and continuity constraints.

\\ \hline
\end{tabular}

\caption{Prompt used to generate candidate storyboard frames using retrieved anchors and global continuity constraints.}

\label{tab:candidate_generation_prompt}

\end{table*}
\begin{algorithm*}[t]
\caption{VLM-Based QA Candidate Selection with Scoring Prompt}
\label{alg:qa_selection}
\small

\textbf{Input:} 
Shot $s_t$, candidate frames $\{I_t^{(1)},...,I_t^{(N)}\}$, global plan $P$, previous frame $I_{t-1}$ \\
\textbf{Output:} Best frame $I_t^*$

\begin{enumerate}

\item Retrieve shot plan from global planner:
\[
p_t = P(s_t)
\]

\item Extract continuity constraints from the global plan:

\begin{itemize}
\item Character appearance states $C(c,t)$
\item Location identity $L(s_t)$
\item Prop states $O(o,t)$
\end{itemize}

\item Construct the \textbf{VLM evaluation prompt} that includes:
\begin{itemize}
\item Shot description $s_t$
\item Character appearance states $C(c,t)$
\item Prop states $O(o,t)$
\item Location identifier $L(s_t)$
\item Previous frame $I_{t-1}$ (if continuation)
\end{itemize}

\item Initialize scores:
\[
score_i = 0 \quad \forall i \in \{1,...,N\}
\]

\item \textbf{Evaluate candidate frames}

For each candidate image $I_t^{(i)}$:

\begin{enumerate}

\item Provide the candidate image and the evaluation prompt to the VLM

\item The VLM evaluates the frame using the scoring rubric and returns:

\begin{itemize}
\item Shot alignment score
\item Character consistency score
\item Background continuity score
\item Prop state correctness score
\end{itemize}

\item Compute the overall score:
\[
score_i = \frac{S_{align} + S_{char} + S_{bg} + S_{prop}}{4}
\]

\end{enumerate}

\item Select best candidate:
\[
I_t^* = \arg\max_i score_i
\]

\item Return selected frame $I_t^*$

\end{enumerate}
\label{algo:vlmqa_asssesement}
\end{algorithm*}

\begin{table*}[t]
\centering
\small

\begin{tabular}{|p{0.95\linewidth}|}
\hline
\rowcolor{blue!15}
\textbf{Prompt: VLM-Based Continuity Scoring Prompt}
\label{tab:qa_prompt} \\ \hline

You are a visual continuity evaluator. Your task is to evaluate a candidate storyboard frame and assign scores based on how well it satisfies the global continuity plan and the current shot description. \\

\textbf{Input}

\begin{itemize}
\item Current shot description $s_t$
\item Candidate image $I_t$
\item Character appearance states from the global plan
\item Prop states for the current shot
\item Location identifier for the scene
\item Previous frame $I_{t-1}$ (if continuation)
\end{itemize}

\textbf{Evaluation Criteria}

Evaluate the image along the following dimensions and assign a score from 1 to 5.

\begin{enumerate}

\item \textbf{Shot Alignment (1–5)}

How well does the image depict the action described in the shot?

\item \textbf{Character Appearance Consistency (1–5)}

Do characters match the planned appearance states?

\item \textbf{Background Continuity (1–5)}

Does the environment match the planned location and maintain spatial continuity with the previous frame?

\item \textbf{Prop State Correctness (1–5)}

Do props appear with the correct state according to the global plan?

\end{enumerate}

\textbf{Return STRICT JSON}

\begin{verbatim}
{
 "shot_alignment": 5,
 "character_consistency": 4,
 "background_continuity": 5,
 "prop_state_correctness": 4,
 "overall_score": 4.5,
 "reasoning": "The frame correctly depicts the museum gallery..."
}
\end{verbatim}

\textbf{Scoring Guidelines}

\begin{itemize}
\item 5 = perfect match with the plan
\item 4 = minor inconsistencies
\item 3 = moderate inconsistencies
\item 2 = major inconsistencies
\item 1 = completely incorrect
\end{itemize}

\\ \hline
\end{tabular}

\caption{Prompt used by the VLM to score candidate storyboard frames during QA-based candidate selection (Algorithm~\ref{alg:qa_selection}).}

\end{table*}

\begin{algorithm*}[t]
\caption{Visual Memory Update with Anchor Extraction}
\label{alg:memory_update}
\small

\textbf{Input:} Selected frame $I_t^*$, shot description $s_t$, global plan $\mathcal{P}$, memory $\mathcal{M}$ \\
\textbf{Output:} Updated memory $\mathcal{M}$

\begin{enumerate}

\item Query the global plan to obtain:
\begin{itemize}
\item Character appearance states $\mathcal{C}(c,t)$
\item Location identity $\mathcal{L}(s_t)$
\item Object states $\mathcal{O}(o,t)$
\end{itemize}

\item Provide the selected frame $I_t^*$ together with the shot description and entity descriptions to a VLM-based extractor (prompt in Table~\ref{tab:memory_prompt}).

\item The VLM identifies visible entities and produces clean visual anchors:
\begin{itemize}
\item Character anchors
\item Location/background anchor
\item Prop anchors
\end{itemize}

\item \textbf{Update Character Memory}

For each detected character $c$:

\[
\mathcal{M}_c(c,\mathcal{C}(c,t)) \leftarrow anchor_c
\]

where $anchor_c$ is the extracted character image representing the current appearance.

\item \textbf{Update Location Memory}

Let $l = \mathcal{L}(s_t)$.

\[
\mathcal{M}_l(l) \leftarrow anchor_l
\]

where $anchor_l$ represents the updated visual environment of the location.

\item \textbf{Update Prop Memory}

For each object $o$ appearing in the frame:

\[
\mathcal{M}_o(o,\mathcal{O}(o,t)) \leftarrow anchor_o
\]

where $anchor_o$ is the extracted visual anchor of the object in its current state.

\item \textbf{Update Frame Memory}

Append the selected frame:

\[
\mathcal{M}_f \leftarrow \mathcal{M}_f \cup \{I_t^*\}
\]

\item Return updated memory $\mathcal{M}$.

\end{enumerate}

\end{algorithm*}

\begin{table}[t]
\centering
\small
\begin{tabular}{|p{0.95\linewidth}|}
\hline
\rowcolor{blue!15}
\textbf{Prompt: Character Anchor Extraction} \\
\hline

You are extracting visual anchors for characters from a storyboard frame.

\textbf{Input}

\begin{itemize}
\item Frame image $I_t$
\item Shot description $s_t$
\item Character appearance states $\mathcal{C}(c,t)$
\end{itemize}

\textbf{Task}

Identify characters visible in the frame and extract a clean visual anchor image that captures their current appearance (face, clothing, hairstyle, and overall identity).

\textbf{Return STRICT JSON}

\begin{verbatim}
{
 "characters":[
   {
    "name":"Ethan",
    "appearance_state":"trench_coat",
    "anchor":"character_anchor_image"
   },
   {
    "name":"Lena",
    "appearance_state":"blue_dress",
    "anchor":"character_anchor_image"
   }
 ]
}
\end{verbatim}

\textbf{Guidelines}

\begin{itemize}
\item Extract anchors only for characters clearly visible.
\item Anchors should capture the full appearance identity.
\end{itemize}

\\
\hline
\end{tabular}

\caption{Prompt for extracting character appearance anchors for updating character memory.}
\label{tab:char_prompt}
\end{table}

\begin{table}[t]
\centering
\small
\begin{tabular}{|p{0.95\linewidth}|}
\hline
\rowcolor{blue!15}
\textbf{Prompt: Location Background Anchor Extraction} \\
\hline

You are extracting the visual environment anchor for a location from a storyboard frame.

\textbf{Input}

\begin{itemize}
\item Frame image $I_t$
\item Shot description $s_t$
\item Location identity $\mathcal{L}(s_t)$
\end{itemize}

\textbf{Task}

Identify the environment corresponding to the location and extract a clean background anchor image that represents the spatial layout of the scene.

\textbf{Return STRICT JSON}

\begin{verbatim}
{
 "location":"museum_gallery",
 "anchor":"background_anchor_image"
}
\end{verbatim}

\textbf{Guidelines}

\begin{itemize}
\item The anchor should capture the environment layout.
\item Avoid including large foreground characters when possible.
\end{itemize}

\\
\hline
\end{tabular}

\caption{Prompt for extracting background anchors associated with locations.}
\label{tab:bg_prompt}
\end{table}

\begin{table}[t]
\centering
\small
\begin{tabular}{|p{0.95\linewidth}|}
\hline
\rowcolor{blue!15}
\textbf{Prompt: Prop Anchor Extraction} \\
\hline

You are extracting visual anchors for important objects from a storyboard frame.

\textbf{Input}

\begin{itemize}
\item Frame image $I_t$
\item Shot description $s_t$
\item Object state descriptions $\mathcal{O}(o,t)$
\end{itemize}

\textbf{Task}

Identify objects that appear in the frame and extract visual anchors representing their current state.

\textbf{Return STRICT JSON}

\begin{verbatim}
{
 "objects":[
   {
    "name":"golden_artifact",
    "state":"inside_display_case",
    "anchor":"prop_anchor_image"
   }
 ]
}
\end{verbatim}

\textbf{Guidelines}

\begin{itemize}
\item Extract anchors only for visible objects.
\item Ensure the object state matches the visual evidence.
\end{itemize}

\\
\hline
\end{tabular}

\caption{Prompt used to extract prop anchors and update object-state memory.}
\label{tab:prop_prompt}
\end{table}

\end{document}